\documentclass[journal]{IEEEtran}
\usepackage{amsfonts}
\usepackage{amsmath}
\usepackage{graphicx}
\usepackage{multirow}
\usepackage{array}
\usepackage{picins}
\usepackage{subfigure}
\usepackage{capt-of}
\usepackage{epsfig}
\usepackage{amssymb}
\usepackage{epstopdf}
\usepackage{setspace}
\usepackage{enumitem}
\usepackage{float}
\usepackage{url}
\usepackage{subfigure}
\usepackage{color}
\usepackage[marginal]{footmisc}

\ifCLASSINFOpdf
\else
\fi
\begin{document}

\title{Global and Local Consistent Wavelet-domain Age Synthesis}

\author{Michael~Shell,~\IEEEmembership{Member,~IEEE,}
        John~Doe,~\IEEEmembership{Fellow,~OSA,}
        and~Jane~Doe,~\IEEEmembership{Life~Fellow,~IEEE}

\author{\IEEEauthorblockN{Peipei Li\IEEEauthorrefmark{2},
Yibo Hu\IEEEauthorrefmark{2},
Ran He \textit{Member, IEEE} and
Zhenan Sun  \textit{Member, IEEE}}

\thanks{\IEEEauthorrefmark{2}These authors contributed to the work equally. }
\thanks{P. Li, Y. Hu, R, He and Z. Sun are with the Center for Research
on Intelligent Perception and Computing, National Laboratory of Pattern
Recognition, Institute of Automation, CAS Center for Excellence in Brain
Science and Intelligence Technology, Chinese Academy of Sciences, 100190,
Beijing, China. E-mail: peipei.li@cripac.ia.ac.cn; yibo.hu@cripac.ia.ac.cn; rhe@nlpr.ia.ac.cn; znsun@nlpr.ia.ac.cn.
}
\thanks{P. Li, R, He and Z. Sun are also with the University of Chinese Academy of
Sciences, Beijing, China. }}
}


\maketitle

\begin{abstract}
Age synthesis is a challenging task due to the complicated and non-linear transformation in human aging process. Aging information is usually reflected in local facial parts, such as wrinkles at the eye corners.
However, these local facial parts contribute less in previous GAN based methods for age synthesis. To address this issue, we propose a Wavelet-domain Global and Local Consistent Age Generative Adversarial Network (WaveletGLCA-GAN), in which one global specific network and three local specific networks are integrated together to capture both global topology information and local texture details of human faces. Different from the most existing methods that modeling age synthesis in image-domain, we adopt wavelet transform to depict the textual information in frequency-domain. 
Moreover, five types of losses are adopted: 1) adversarial loss aims to generate realistic wavelets; 2) identity preserving loss aims to better preserve identity information; 3) age preserving loss aims to enhance the accuracy of age synthesis; 4) pixel-wise loss aims to preserve the background information of the input face; 5) the total variation regularization aims to remove ghosting artifacts.
Our method is evaluated on three face aging datasets, including CACD2000, Morph and FG-NET. Qualitative and quantitative experiments show the superiority of the proposed method over other state-of-the-arts.
\end{abstract}

\begin{IEEEkeywords}
Age synthesis, wavelet transform, generative adversarial network, global and local features.
\end{IEEEkeywords}

\IEEEpeerreviewmaketitle

\section{Introduction}
\IEEEPARstart{H}{uman} age synthesis, also known as age progression/regression, has captured much attention in many real-world applications such as finding missing children, age estimation, cross-age face verification, social entertainment, etc. It aims to aesthetically rerender a given face with natural aging effects but still preserve personality. Impressive progress has been made on age synthesis in recent years and many methods \cite{zhang2017age,yang2017learning,duong2017temporal,zhou2017personalized,wang2018face} have been proposed. However, large-scale age synthesis is still a very challenging problem for several reasons. 1) Rigid requirement for the training and testing datasets, i.e., most existing works require face images of the same subject at different ages. 2) Faces captured in the wild are usually influenced by the large variations of illumination, pose, resolution and expression. 3)
Different subjects have different aging processes. There are many internal and external influential factors, e.g., gene, gender, living style.

In prior works, physical modeling based methods and prototype-based methods are mainly used for age synthesis. Physical modeling based methods build computational models for age progression/regression to simulate human biological and aging mechanisms of the cranium, muscles or facial skin etc. Although rich and obvious textures can be generated, these approaches depend on many face sequences of the same subject covering a long age span, which is hard and expensive to obtain. Besides, in prototype-based methods, only texture synthesis is considered. Training data is divided into different age groups. Average face of each age group is regarded as its prototype, thus the aging pattern is the difference between prototypes of target and source age groups. However, prototype-based approaches may ignore identity and different subjects share the same aging pattern.
\begin{figure}[t]
\setlength{\abovecaptionskip}{0cm}
\begin{center}
\includegraphics[width=0.88\linewidth]{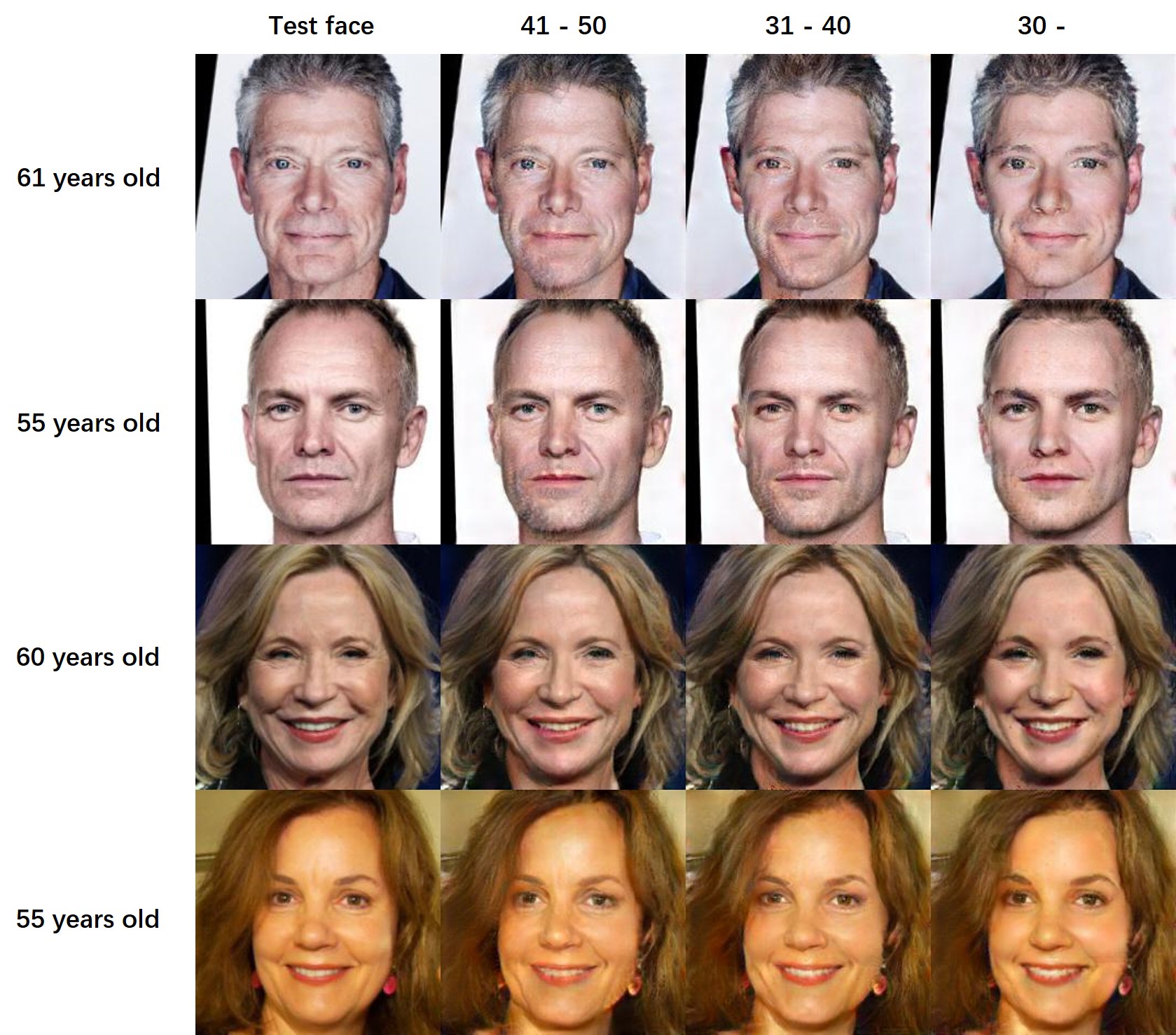}
\end{center}

   \caption{Demonstration of our age regression results. For each subject, the first column is the input, while the rest three columns are the rejuvenating results from older to younger. (Zoom in for a better view.) }\label{fig:illu}

\end{figure}
Recently, deep learning based methods have made a breakthrough in computer vision. Training with large-scale data, deep neural networks substantially show better performance than previous methods in image generation. Generative Adversarial Networks (GANs) \cite{goodfellow2014generative} as a specific type of deep neural network, has proven capable of generating photo-realistic images and is therefore introduced to deal with age synthesis \cite{nhan2016longitudinal,wang2016recurrent,zhang2017age,yang2017learning,duong2017temporal,zhou2017personalized, wang2018face}.
However, the generated results of existing GAN based methods are often over-smoothed and lack of detailed texture information. Fine-grained age synthesis remains an open and challenging task.

We tackle this fine-grained age synthesis from two aspects: modeling aging patterns of age-sensitive areas with local networks and introducing wavelet transform to age synthesis. In the first place, aging texture information is usually reflected in the local facial parts, such as wrinkles at the eye corners or on the forehead. However, existing GAN based models cannot fully capture these rich textures in the local facial parts. Secondly, most current age synthesis methods only depend on modeling aging process in image-domain. Thus the outputs of these methods tend to be over-smoothed and lack of textural details. Since the subtle texture information is more salient and robust in frequency-domain, we introduce wavelet transform to age synthesis.
\begin{figure}[t]
\setlength{\abovecaptionskip}{0cm}
\begin{center}
\includegraphics[width=1\linewidth]{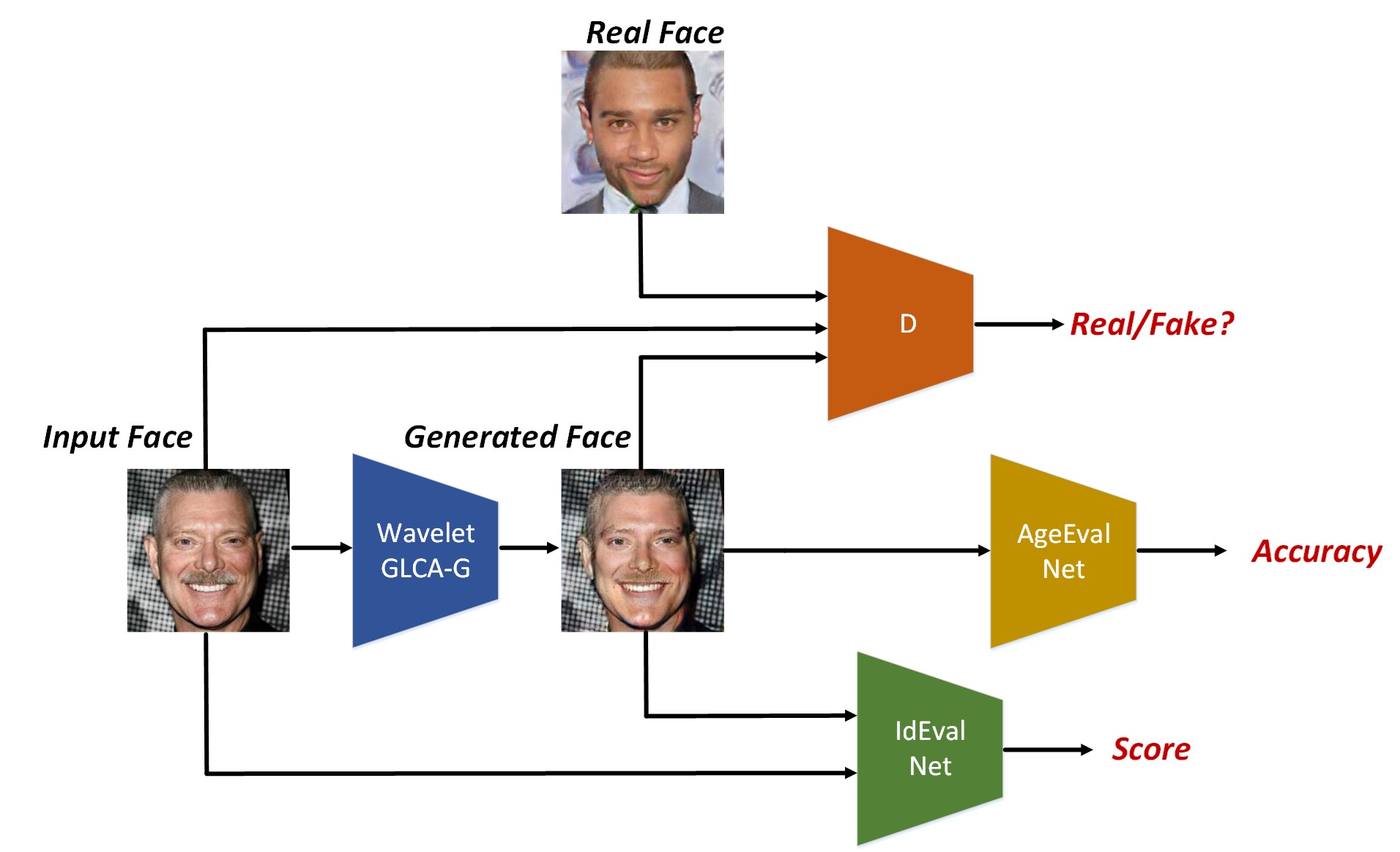}
\end{center}
\vspace{-0.4cm}
   \caption{General framework of WaveletGLCA-GAN. }\label{fig:fram}
\vspace{-0.5cm}
\end{figure}

In this paper, a novel Wavelet-domain Global and Local Consistent Age Generative Adversarial Network (WaveletGLCA-GAN) is proposed for fine-grained age synthesis. It consists of four parts: wavelet-domain global and local consistent age generator, discriminator, age evaluation network, identity evaluation network. Fig. \ref{fig:illu} and Fig. \ref{fig:fram} depict the rejuvenating results and framework of WaveletGLCA-GAN, respectively.

The wavelet-domain global and local consistent age generator (WaveletGLCA-G) takes a random age face as an input and predicts the corresponding wavelet coefficients before reconstructing the target age output. Apart from one global specific network like previous works, three local specific networks are imposed to capture the aging texture information in forehead, eyes and mouth parts, respectively. Besides, the wavelet coefficient prediction network is employed to predict wavelet coefficients of the target age faces. Instead of manipulating a whole face, our WaveletGLCA-G learns the residual face, which can accelerate the convergence while preserving the identity and background information.
The discriminator is proposed to harness target age information and tries to distinguish the synthesized faces from the real target age faces. Through adversarial training, the WaveletGLCA-G is forced to produce photo-realistic target age images.

Aging accuracy is one of the two critical requirements in age synthesis. Thus, we adopt the age evaluation network to evaluate and reinforce the aging accuracy in age synthesis. The age evaluation network takes the generated face as an input and estimates its age.
Meanwhile, identity preserving performance is the other critical requirement in age synthesis. An identity evaluation network is introduced to evaluate and enhance the identity preserving performance in age synthesis. It takes the generated and participant faces as inputs and computes their similarity scores to evaluate how much identity information can be retained.

The main contributions of this work are as follows:

1)	A global and local consistent age generator is proposed for deep facial age synthesis, in which three local specific networks and one global specific network are integrated together to capture both global topology and local texture information of human faces.

2)  Different from the most existing studies that model age synthesis in image-domain, we transform age synthesis to a wavelet coefficient prediction task in frequency-domain. To the best of our knowledge, this is the first attempt to transform age synthesis to wavelet coefficient prediction task in GAN-based framework for age synthesis.




3)	The proposed WaveletGLCA-GAN simultaneously achieves age progression and regression in the same framework with the given age labels. Experimental results show that the proposed approach outperforms state-of-the-art methods in terms of qualitative and quantitative metrics.

This paper is an extension of our previous method GLCA-GAN \cite{li2018global}. Apart from providing more in-depth analysis and more extensive experiments, the major difference between this paper and its previous version lies in three-folds: 1) Age synthesis is extended from image-domain generative adversarial network (GLCA-GAN) to wavelet-domain generative adversarial network (WaveletGLCA-GAN) with the impose of frequency-domain information which is more sensitive to texture details. 2) Age synthesis is transformed into wavelet coefficients prediction task. Wavelet coefficient prediction network as well as wavelet reconstruction networks are used to capture and model the delicate texture information in wavelet-domain. Besides, higher-resolution age synthesis is achieved with pixel-size extended from $128 \times 128$ to $224 \times 224$. 3) Age estimation and identity verification are proposed to evaluate the aging accuracy and identity preserving performance of WaveletGLCA-GAN.

\begin{figure*}[t]
\setlength{\abovecaptionskip}{0cm}
\begin{center}
\includegraphics[width=1\linewidth]{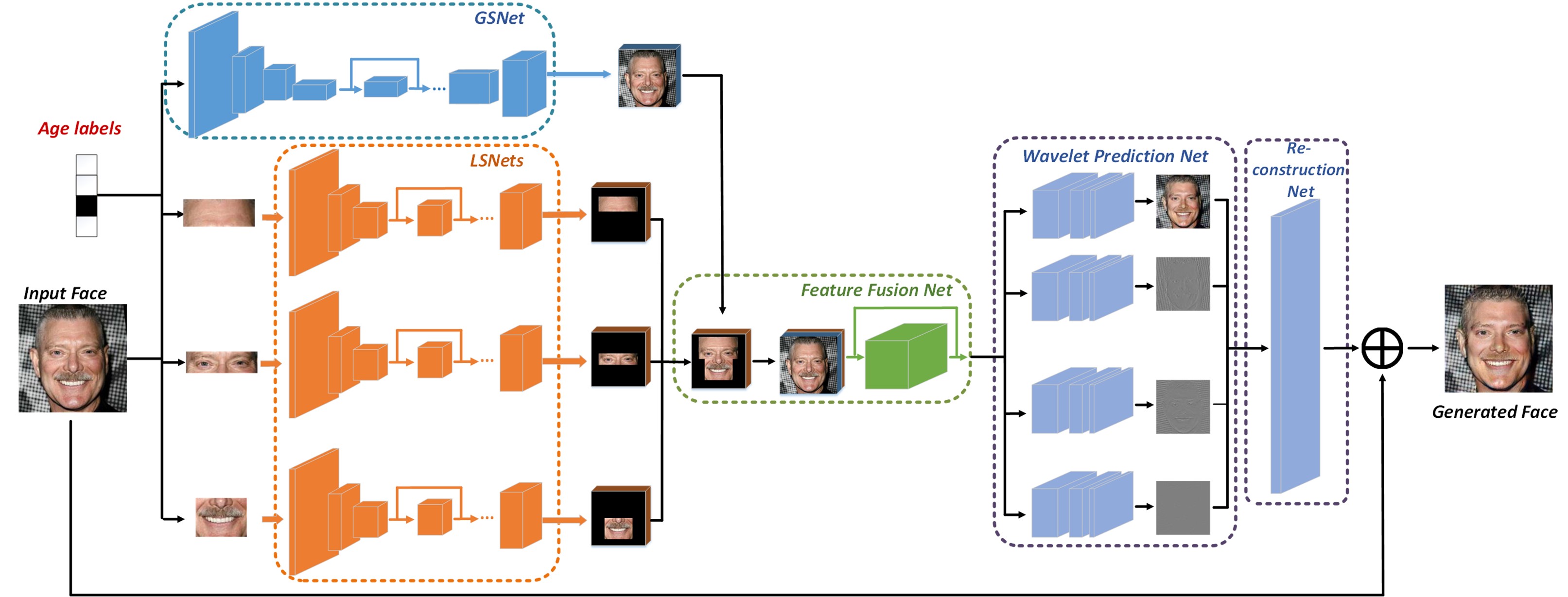}
\end{center}

\vspace{-0.2cm}
   \caption{General framework of WaveletGLCA-G, including global specific network, local specific networks, feature fusion network, wavelet coefficient prediction network as well as wavelet reconstruction network. }\label{Fig:G}
\vspace{-0.2cm}
\end{figure*}

\section{Related Work}
\subsection{Image-to-image Translation}
Age synthesis is a subproblem of image-to-image translation \cite{huang2017wavelet,liu2017unsupervised,liu2016coupled,huang2018introvae} which transforms the input image from the source domain to the target domain with the original structure or semantics preserving. In recent years, image translation has been widely investigated. A host of recent works have explored the ability of Generative Adversarial Networks (GANs) \cite{goodfellow2014generative} for image-to-image translation. With a generator and a discriminator, GANs play a min-max game to learn the target image prior.
pix2pix \cite{Isola2016Image} learns image-to-image translation in a supervised manner using cGANs \cite{Mirza2014Conditional}, which requires paired data samples. StarGAN \cite{Choi2017StarGAN} adds a mask vector to the domain label, which enables joint training between multiple domains.
To alleviate the requirement to paired data, unpaired image-to-image translation has been proposed. Cycle-consistent Adversarial Networks (CycleGANs) \cite{zhu2017unpaired} is successfully applied for unpair image-to-image translation. Recently a Couple-Agent Pose-Guided Generative Adversarial Network (CAPG-GAN) \cite{hu2018pose} is proposed for arbitrary view face synthesis which combines prior domain knowledge of pose and local structure of face to reinforce the realism of results. \cite{chang2018pairedcyclegan} introduces the idea of asymmetric style transfer and proposes PairedCycleGAN, which involves two asymmetric networks for makeup transfer and removal.
\subsection{Age Synthesis}
The facial age synthesis methods can be mainly classified into three categories: physical modeling based, prototype based and deep learning based methods.

In the early time, physical modeling based methods and prototype based methods are mainly adopted in age synthesis. The physical modeling based methods \cite{tsai2014human, suo2010compositional, todd1980perception, suo2012concatenational, ramanathan2006modeling} simulate age progression/regression by modeling human biological and physical mechanisms, e.g., muscles, facial contours. However, this kind of approaches need abundant face sequences of the same subject covering a long age span, which is expensive and hard to obtain. Besides, prototype-based approaches \cite{tiddeman2001prototyping, kemelmacher2014illumination} regard the average face of each age group as its prototype, thus the aging pattern is the difference between the prototypes of target and source age groups. However, prototype-based approaches ignore identity information. To handle this problem, Shu et al. \cite{shu2015personalized} propose an age synthesis method based on dictionary learning to reconstruct the aging face.

Recently, deep learning based approaches \cite{nhan2016longitudinal, wang2016recurrent, zhang2017age, yang2017learning, duong2017temporal, zhou2017personalized, wang2018face} have shown considerable ability in age synthesis. Wang et al. \cite{wang2016recurrent} propose a recurrent neural network to model a smoother face aging process. Zhang et al. \cite{zhang2017age} apply Conditional Adversarial Autoencoder (CAAE) to synthesize target age faces with target age labels. In addition, Zhou et al. \cite{zhou2017personalized} argue that occupation information influences the individual aging process and propose an occupational-aware adversarial face aging network. To make the most of the discriminative ability of GAN, Yang et al. \cite{yang2017learning} put forward a multi-pathway discriminator to refine the aging/rejuvenating results. Duong et al. \cite{duong2017temporal} present a generative probabilistic model to simulate the aging mechanism of each age stage. Wang et al. \cite{wang2018face} impose an identity-preserved term and an age classification term into age synthesis and propose the Identity-Preserved Conditional Generative Adversarial Networks (IPCGANs). However, the age synthesis results of existing methods are still over-smoothed and lack of detailed texture information, which hurts the aging accuracy and identity preserving performance.

\section{Approach}
Age synthesis aims to synthesize a target age face image $\hat I^t$ from a given age face image $I^g$. Our goal is to learn such a synthesizer that can infer the corresponding age images. The overall framework of our proposed Wavelet-domain Global and Local Consistent Age Generative Adversarial Network (WaveletGLCA-GAN) is depicted in Fig. 2, which mainly consists of four parts: wavelet-domain global and local consistent age generator, discriminator, age evaluation network, identity evaluation network.

In WaveletGLCA-GAN, age information is encoded into a one-hot vector $l$ with the target age being 1. In this paper we divide the ages into four age groups, thus the age vector ${l} \in {R^{1 \times 4}}$. We stack $l$ to ${L} \in {R^{H \times W \times 4}}$ as the input age label, where $H$ and $W$ are the height and width of the input image, respectively. Given an input age face $I^g$, forehead, eyes and mouth local patches $p_1$, $p_2$, $p_3$ are first cropped according to the five facial landmarks. Then the age label $L$ is concatenated to both $I^g$ and three local facial patches $p_1$, $p_2$, $p_3$. The new global tensor $\left[ {I^g,L} \right]$ and three local tensor $\left[{p_1,L} \right]$, $\left[ {p_2,L} \right]$, $\left[ {p_3,L} \right]$ are fed into WaveletGLCA-G to sythesize the target age image $\hat I^t$.
Five types of losses are adopted to supervise the WaveletGLCA-G to achieve accurate age generation under the premise of preserving the background and identity information.


This section firstly introduces the wavelet transform and the architecture of the WaveletGLCA-G, discriminator, age evaluation network and identity evaluation network. Then the details of five supervised loss functions, including conditional adversarial loss, identity preserving loss, age preserving loss, pixel-wise loss and total variation regularization, are presented later.
\begin{figure}[t]
\setlength{\abovecaptionskip}{0cm}
\begin{center}
\includegraphics[width=1\linewidth]{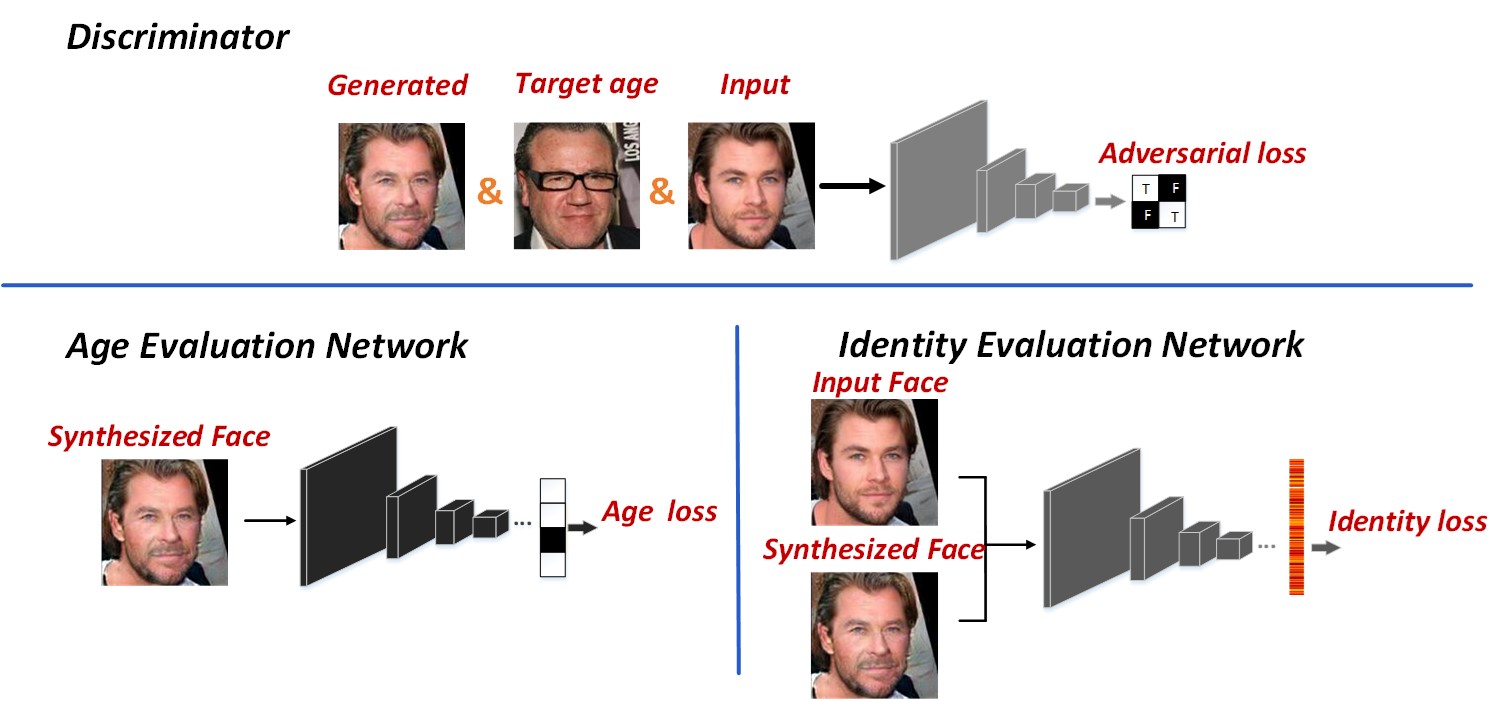}
\end{center}
\vspace{-0.2cm}
   \caption{General framework of constraint networks, including discriminator, age evaluation network, identity evaluation network.}\label{fig:constraint}
\vspace{-0.4cm}
\end{figure}

\subsection{Wavelet Transform}
Most current age synthesis methods only model aging process in image-domain, whose synthesized results usually tend to be over-smoothed and lack of textural details. To alleviate this limitation, we employ wavelet packet transform, which decomposes the image into the approximation (low-frequency component) and the details (high-frequency components). In this paper, the low-frequency component is related to the general face information, while the high-frequency components are related to the detailed aging texture information.

\begin{figure}[t]
\setlength{\abovecaptionskip}{0cm}
\begin{center}
\includegraphics[width=1\linewidth]{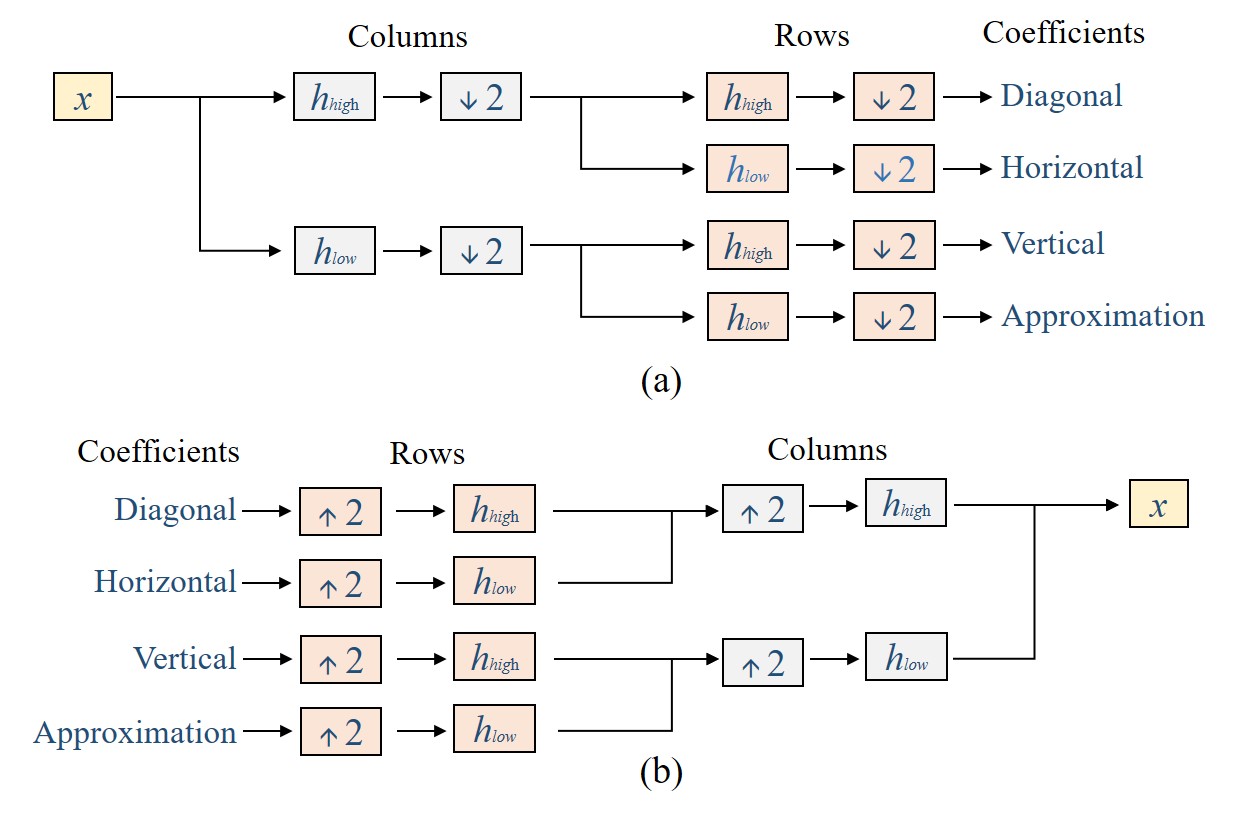}
\end{center}
\vspace{-0.2cm}
   \caption{Illustration of wavelet decomposition and reconstruction in fast wavelet transform (FWT). (a) is the process of wavelet packet decomposition, where a cascade of filters with $h_{high}$ and $h_{low}$ followed by a factor 2 subsampling are used to compute the wavelet coefficients. (b) is the process of wavelet packet reconstruction, which does the inverse operation of (a).}\label{fig:FWT}
\vspace{-0.2cm}
\end{figure}

Fig. \ref{fig:FWT} depicts the process of wavelet decomposition and reconstruction. As shown in Fig. \ref{fig:FWT}(a), the original image is firstly decomposed into an approximation coefficient and three detailed coefficients. Then according to the wavelet components, the image is reconstructed as shown in Fig. \ref{fig:FWT}(b). Inspired by this reversible process, we propose this wavelet-domain age synthesis method.

Note that in the proposed method, the wavelet coefficients are not explicitly decomposed from the original image by wavelet decomposition, but are directly learned from the global and local fused features. Specifically, we initialize the wavelet reconstruction network with a pre-computed wavelet reconstruction matrix and fix its parameters during training, which restrains the wavelet prediction network to directly learn the four corresponding wavelet coefficients from the fused features.

\subsection{Network Architecture}

\subsubsection{\textbf{Architecture of WaveletGLCA-G}}
The wavelet-domain global and local consistent age generator (WaveletGLCA-G) takes the source age faces as inputs and synthesize the corresponding target age faces. As shown in Fig. \ref{Fig:G}, our WaveletGLCA-G consists of five subnetworks: global specific network, local specific networks, feature fusion network, wavelet coefficient prediction network as well as wavelet reconstruction network. The GSNet and three LSNets learn the feature embeddings of the whole facial topology and the subtle textures of forehead, eyes and mouth simultaneously. Then the feature fusion network fuses the four feature embeddings of the global and local networks. After that, the wavelet coefficient prediction network estimates the  wavelet coefficients by four individual subnetworks. Finally, according to the predicted coefficients, the wavelet reconstruction network reconstructs the target age face.
Instead of manipulating a whole face, our WaveletGLCA-G learns the residual face, which accelerates the convergence of WaveletGLCA-GAN while preserving the identity and background information.

As shown in Fig. \ref{Fig:G}, we concatenate the given image $I^g$, the target age label $L$ as the input of WaveletGLCA-G and synthesize the target age image $\hat I^t$.
\begin{equation}
\begin{array}{c}
$${\hat I^t} = {G_{{\theta _G}}}({I^g},{L})$$
\end{array}
\end{equation}
where ${\theta _G}$ is the parameter of WaveletGLCA-G. In this paper, we choose the image size of 224 and divide the ages of subjects into four age groups, thus ${L} \in {R^{224 \times 224 \times 4}}$. The detailed architecture of WaveletGLCA-G is shown in Table \ref{tab:G}.
\begin{table}[htbp]\scriptsize
\centering
\caption{the architecture of WaveletGLCA-G $G_{\theta _G}$.}
\begin{tabular}{l|l|l|l} \hline
Layer & Input  & Filter/Stride & Output Size\\
\hline
\multicolumn{4}{c}{GSNet} \\
\hline
G$\_$conv1 & $I^g$, $l^t$         & $5\times5 / 1$ & $224 \times 224 \times 64$\\
G$\_$conv2 & G$\_$conv1                & $3\times3 / 2$ & $112 \times 112 \times 128$\\
G$\_$conv3 & G$\_$conv2                & $3\times3 / 2$ & $56 \times 56 \times 256$\\
G$\_$conv4 & G$\_$conv3                & $3\times3 / 2$ & $28 \times 28 \times 512$\\
G$\_$res1  & G$\_$conv4                & $3\times3 / 1$ & $28 \times 28 \times 512$\\
G$\_$res2  & G$\_$res1                 & $3\times3 / 1$ & $28 \times 28 \times 512$\\
G$\_$deconv1 & G$\_$res2                & $3\times3 / 2$ & $56 \times 56 \times 256$\\
G$\_$deconv2 & G$\_$deconv1                & $3\times3 / 2$ & $112 \times 112 \times 256$\\
\hline
\multicolumn{4}{c}{LSNet} \\
\hline
L$\_$conv1 & $p_1/p_2/p_3$, $l^t$         & $5\times5 / 1$ & $224 \times 224 \times 64$\\
L$\_$conv2 & L$\_$conv1                & $3\times3 / 2$ & $112 \times 112 \times 128$\\
L$\_$conv3 & L$\_$conv2                & $3\times3 / 2$ & $56 \times 56 \times 256$\\
L$\_$res1 & L$\_$conv3                 & $3\times3 / 1$ & $56 \times 56 \times 256$\\
L$\_$res2       & L$\_$res1            & $3\times3 / 1$ & $56 \times 56 \times 256$\\
L$\_$deconv1 & L$\_$res2                & $3\times3 / 2$ & $112 \times 112 \times 256$\\
\hline
\multicolumn{4}{c}{Feature Fusion Net} \\
\hline
F$\_$res1 & G$\_$deconv2, L$\_$deconv1         & $3\times3 / 1$ & $112 \times 112 \times 512$\\
F$\_$res2 & F$\_$res1                & $3\times3 / 1$ & $112 \times 112 \times 512$\\
\hline
\multicolumn{4}{c}{Wavelet Prediction Net} \\
\hline
W$\_$conv1 & F$\_$res2         & $3\times3 / 1$ & $112 \times 112 \times 256$\\
W$\_$conv2 & W$\_$conv1                & $3\times3 / 1$ & $112 \times 112 \times 128$\\
W$\_$conv3 & W$\_$conv2                & $3\times3 / 1$ & $112 \times 112 \times 64$\\
W$\_$conv4 & W$\_$conv3                & $3\times3 / 1$ & $112 \times 112 \times 3$\\
\hline
\multicolumn{4}{c}{Reconstruction Net} \\
\hline
R$\_$deconv1 & W$\_$conv4         & $2\times2 / 2$ & $224 \times 224 \times 3$\\
\hline

\end{tabular}\label{tab:G}
\end{table}

Specifically, the global specific network $G_{g}$ takes the whole image ${I^g} \in {R^{224 \times 224 \times 3}}$ concatenated with the target age label ${L}$ as the input $\left( {{I^g},L} \right)\in {R^{224 \times 224 \times 7}}$. An encoder-decoder architecture with four strided convolutional layers, two residual blocks and two deconvolutional layers are exploited to extract and transform the global feature maps $G_{g}(I^g,L) \in {R^{112 \times 112 \times 256}}$.
In addition, the three local specific networks $G_{l}^i,i \in \left\{ {1,2,3} \right\}$ take forehead, eyes and mouth local patches $p_i$ concatenated with the target age label ${L}$ as the inputs and learn a separate set of filters to extract and transform the corresponding local feature maps $G_{l}^i(p_i,L) \in {R^{112 \times 112 \times 256}}$. All the local specific networks adopt the encoder-decoder architecture with three strided convolutional layers, two residual blocks and one deconvolutional layer.

The feature fusion network fuses the four feature embeddings of global and local networks by two residual blocks.
The wavelet coefficient prediction network has four parallel independent subnetworks. Each of the subnetworks takes the whole output of the fusion network as its input and predicts the corresponding wavelet coefficients, including the approximation coefficient, the detail horizontal, vertical and diagonal coefficients. All the predicted coefficients are half the size of the input, i.e, $112 \times 112 \times 3$.

Finally, the reconstruction network takes the four predicted wavelet coefficients as its input and  reconstructs the target age face $\hat I^t\in {R^{224 \times 224 \times 3}}$. The reconstruction network is realized by a deconvolutional layer, which is initialized by a constant wavelet reconstruction matrix. The parameters of the reconstruction network are pre-computed and can be downloaded from the released code$\footnote{\url{https://github.com/hhb072/WaveletSRNet}}$. Note that, during training, the parameters of reconstruction network are fixed. In this way, the gradients from the age evaluation and identity evaluation networks can be propagated backwards, which leads to an end-to-end training of the proposed WaveletGLCA-G.

\subsubsection{\textbf{Architecture of Discriminator}}
To exploit the prior domain knowledge of ages, we impose age labels into the discriminator $D_{{\theta _D}}$ to further force the generation of age-specific faces. ${\theta _D}$ is the parameter of the proposed discriminator. Particularly, both the input face $I^g$ and the synthetic face $\hat I^t$ with the target age label $L$ are treated as negative samples, while the real face $I^t$ with $L$ is the positive sample. The discriminator takes them as the inputs and outputs a scalar, representing the probability that the inputs come from the real data. Based on the GAN principle, our discriminator $D_{{\theta _D}}$ forces the WaveletGLCA-G to synthesize realistic and plausible target age faces, which are indistinguishable with the real faces. The detailed architecture of our discriminator is shown in Table \ref{tab:D}.
\begin{table}[htbp]\scriptsize
\centering
\caption{The architecture of Discriminator $D_{\theta _D}$.}
\begin{tabular}{l|l|l|l} \hline
Layer & Input  & Filter/Stride & Output Size\\
\hline

D$\_$conv1 & $I^g$,$I^t$,$\hat I^t$,$l^t$         & $4\times4 / 2$ & $112 \times 112 \times 32$\\
D$\_$res1 & D$\_$conv1                & $3\times3 / 1$ & $112 \times 112 \times 64$\\

D$\_$conv2 & D$\_$res1                & $4\times4 / 2$ & $56 \times 56 \times 64$\\
D$\_$res2 & D$\_$conv2                & $3\times3 / 1$ & $56 \times 56 \times 64$\\

D$\_$conv3 & D$\_$res2                & $4\times4 / 2$ & $28 \times 28 \times 128$\\
D$\_$res3 & D$\_$conv3                & $3\times3 / 1$ & $28 \times 28 \times 128$\\

D$\_$conv4  & D$\_$res3                & $4\times4 / 2$ & $14 \times 14 \times 256$\\
D$\_$res4  & D$\_$conv4                 & $3\times3 / 1$ & $14 \times 14 \times 256$\\

D$\_$conv5  & D$\_$res4                & $4\times4 / 2$ & $7 \times 7 \times 256$\\
D$\_$res5  & D$\_$conv5                 & $3\times3 / 1$ & $7 \times 7 \times 256$\\

D$\_$conv5  & D$\_$res4                & $4\times4 / 1$ & $6 \times 6 \times 1$\\
\hline

\end{tabular}\label{tab:D}
\end{table}

\subsubsection{\textbf{Architecture of Identity Evaluation Network}}

 An identity evaluation network is employed to preserve the identity information in age synthesis. It takes the generated face $\hat{I^t}$ and the input face $I^g$ as the inputs and computes their similarity scores to evaluate how much identity information is preserved. We choose a pretrained Light CNN \cite{wu2015light} as the identity evaluation network and fix the parameters during training procedure.
\subsubsection{\textbf{Architecture of Age Evaluation Network}}
An age evaluation network is introduced in WaveletGLCA-GAN to reinforce the aging accuracy in age synthesis. It takes the generated face $\hat{I^t}$ as the input and estimate the corresponding age to evaluate how much age-related information has been learned. We choose a pretrained VGG16 \cite{simonyan2014very} as the age evaluation network and fix the parameters during training procedure.

\subsection{Training Losses}
A weighted sum of five losses is imposed to supervise the proposed WaveletGLCA-GAN in an end-to-end manner, including conditional adversarial loss, identity preserving loss, age preserving loss, pixel-wise loss and total variation regularization.

\subsubsection{\textbf{Conditional Adversarial Loss}}
To make the synthesized images indistinguishable from the real data and incorporate prior domain knowledge (age information from target age group), the adversarial loss conditioned on the target age label is introduced to our WaveletGLCA-GAN. In addition, for avoiding producing artifacts as in \cite{shrivastava2016learning}, our discriminator $D_{{\theta _D}}$ distinguishes the $6 \times 6$ local image patches separately. The adversarial loss is formulated as:

\begin{equation}
\begin{array}{c}
$$\begin{array}{l}{L_{adv}} = \mathop {\min\limits_{\theta _G}} {\max\limits_{\theta _D}} {E_{{I^t},L{\sim }p({I^t},L)}}[\log D_{{\theta _D}}({I^l},L)] + \\\hspace{0.5cm} \;\;\;\;\;\;\;{E_{{I^g},L{\sim }p({I^g},L)}}[\log (1 - D_{{\theta _D}}({I^g},L))] + \\\;\;\;\;\;\;\;\;\;\;\;\;{E_{{I^g},L{\sim }p({I^g},L)}}[\log (1 - D_{{\theta _D}}({G_{{\theta _G}}}({I^g},L),L))]\vspace{0.1cm}\end{array}

$$
\end{array}
\end{equation}
where $I^t$ denotes the real face of age group $L$. Parameters of the generator and the discriminator are trained alternately to optimize the min-max problem.
\subsubsection{\textbf{Identity Preserving Loss}}
It is crucial to keep identity information in the age synthesis. However, the synthetic faces based on GAN are usually close to the real data only in pixel space, not in semantic space.  Hence, we introduce an identity preserving loss to our model. Following \cite{li2018Prot}, the identity preserving loss is formulated as:
\begin{equation}
\begin{array}{c}
$$\begin{array}{l}{L_{id}} = ||{\varphi _f}({I^g}) - {\varphi _f}({G_{{\theta _G}}}({I^g},L))||_2^2 + \\\;\;\;\;\;\;\;\;\;\;||{\varphi _p}({I^g}) - {\varphi _f}({G_{\theta p}}({I^g},L))|{|_F}\end{array}

$$\vspace{0.1cm}
\end{array}
\end{equation}
where ${\varphi _f}$ and ${\varphi _p}$ denote the feature extractors of the fully-connected layer and the last pooling layer of the pre-trained light CNN-29 network \cite{wu2015light} respectively.

\subsubsection{\textbf{Age Preserving Loss}}
Age accuracy is another key issue of age progression/regression. In our model, an age preserving loss based on the pre-trained VGG16 structure is utilized to enhance the age accuracy of the synthetic face, which can be formulated as:
\begin{equation}
\begin{array}{c}
{L_{age}} =  - \frac{1}{N}\sum\limits_{i = 1}^N {\sum\limits_{j = 0}^3 {\left( {1\left\{ {{y^i} = j} \right\}\log \left( s \right)} \right)} }
\end{array}
\end{equation}
where $s$ denotes the output of the final softmax layer of VGG16, $y^i$ denotes the age class of the i-th synthetic face. When $y^i$ is equal to the $j$, $1\left\{ {{y^i} = j} \right\} = 1$, otherwise $1\left\{ {{y^i} = j} \right\} = 0$.
\subsubsection{\textbf{Pixel-wise Loss}}
By minimizing the adversarial, identity preserving and age preserving losses, $G_{{\theta _G}}$ can generate photo-realistic faces with accurate identity and age information. However, minimizing these losses can't guarantee the background well preserved. Since there is no ground-truth target face in age synthesis, a pixel-wise $L2$ loss between the synthetic and the input faces is adopted to preserve the background.
\begin{equation}
\begin{array}{c}
$${L_{pixel}} = \frac{1}{{C \times H \times W}}||{I^g} - {G_{{\theta _G}}}({I^g},L)||_2^2$$
\end{array}
\end{equation}
where $C$, $H$, $W$ are the channel, height and width of the image respectively. It is worth noting that Eq. (5) forces the synthetic face image to be similar with the input face image. Thus we update this pixel-wise loss at every 10 iterations to balance the age synthesis and the background preservation.

\subsubsection{\textbf{Total Variation Regularization}}
The images generated by GAN usually have some ghosting artifacts \cite{radford2015unsupervised}, which deteriorates visualization as well as recognition performance. A total variation regularization term is thus imposed to remove the unfavorable artifacts.
\begin{equation}
\begin{array}{c}
$${L_{tv}} = \sum\limits_{c = 1}^C {\sum\limits_{w,h = 1}^{W,H} {|\hat I_{w + 1,h,c}^t - \hat I_{w,h,c}^t|} }  + |\hat I_{w,h + 1,c}^t - \hat I_{w,h,c}^t|

$$
\end{array}
\end{equation}

\subsubsection{\textbf{Overall Loss}}
Finally, the total supervised loss is a weighted sum of the above five losses. ${G_{{\theta _G}}}$ and $D_{{\theta _D}}$ are trained alternately to optimize the min-max problem. Taking all loss functions together, the overall loss can be written as:
\begin{equation}
\begin{array}{c}
{L} = {\lambda _1}{L_{adv}} + {\lambda _2}{L_{ip}} + {\lambda _3}{L_{age}} + {\lambda _4}{L_{pixel}}+{\lambda _5}{L_{tv}}
\end{array}
\end{equation}
where ${\lambda _1}$, ${\lambda _2}$, ${\lambda _3}$, ${\lambda _4}$, ${\lambda _5}$ are trade-off parameters that control the relative importance among conditional adversarial loss, identity preserving loss, age preserving loss, pixel-wise loss and total variation regularization.
\section{Experiments}
WaveletGLCA-GAN provides a flexible way to rerender the input face to any age group controlled by the age label. It produces photo-realistic faces with accurate identity and age information. In the following subsections, we begin with an introduction of datasets and settings. Then we demonstrate the superiority of our WaveletGLCA-GAN on both qualitative visualization  and quantitative face verification and age estimation. Lastly, we conduct ablation study to demonstrate the benefits gained from local specific networks and wavelet coefficient prediction network.

\begin{figure*}[t]
\setlength{\abovecaptionskip}{0cm}
\begin{center}
\includegraphics[width=1\linewidth]{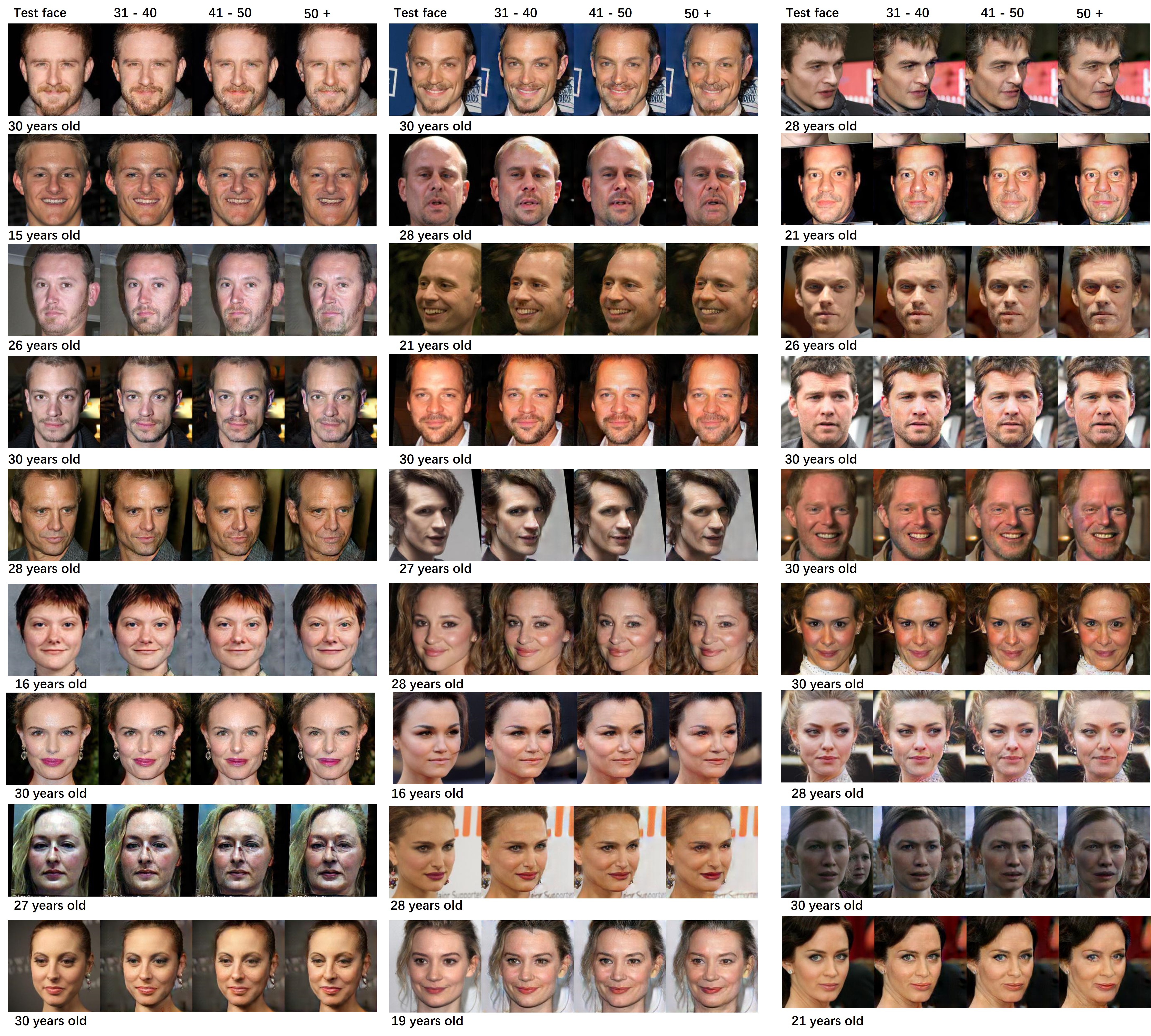}
\end{center}
\vspace{-0.3cm}
   \caption{Age progression results on the CACD2000 dataset for 27 different subjects. For each subject, the leftmost column shows the input face, while the rest three columns are synthetic faces from younger to older. (Zoom in for a better view.)}
\vspace{-0.3cm}
\label{fig:cacd_progression}
\end{figure*}
\begin{figure*}[t]
\setlength{\abovecaptionskip}{0cm}
\begin{center}
\includegraphics[width=1\linewidth]{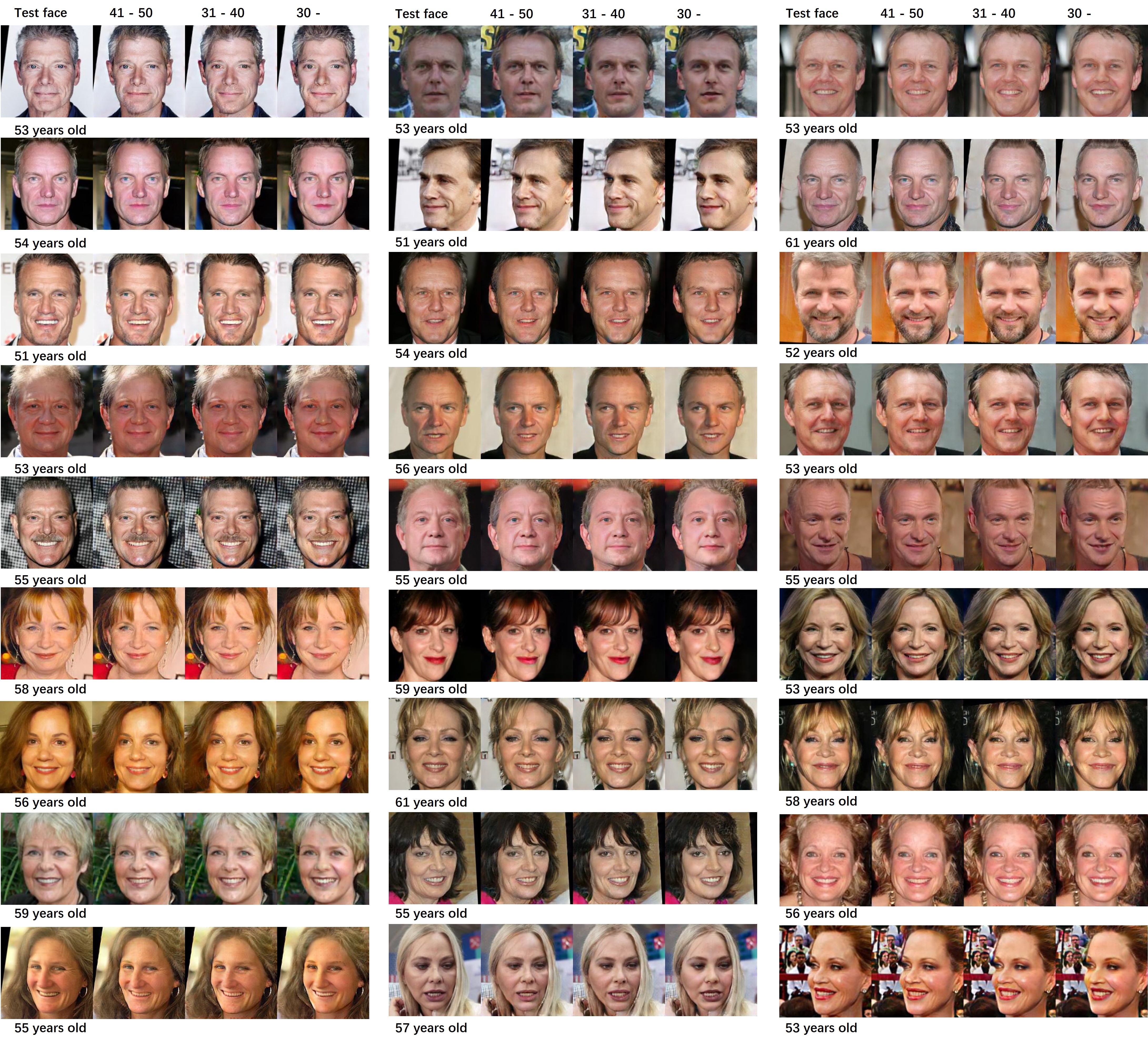}
\end{center}
\vspace{-0.3cm}
   \caption{Age regression results on the CACD2000 dataset for 27 different subjects. For each subject, the leftmost column shows the input face, while the rest three columns are synthetic faces from older to younger. (Zoom in for a better view.)}
\vspace{-0.3cm}
\label{fig:cacd_regression}
\end{figure*}

\subsection{Datasets and Settings}

\subsubsection{\textbf{Datasets}}
We employ three datasets, including CACD2000 \cite{chen2015face}, Morph \cite{ricanek2006morph} and FG-NET \cite{lanitis2002toward}, to evaluate the proposed WaveletGLCA-GAN.

\begin{itemize}[itemindent=1em]
\item \textbf{CACD2000} \cite{chen2015face}: The CACD2000 dataset contains 163,446 color images of 2,000 celebrities with aging range from 14 to 62 years old. It has more than 80 images for per subject on average. The maximum age gap between images for the same subject is 10 years. With variations of age in the wild, it has been widely used to evaluate age synthesis and age invariant verification performance under unconstrained environments. Since the face images in CACD2000 are collected from the web and contain variations in illumination, pose, expression, etc., it is extremely challenging for age synthesis task.
\item \textbf{Morph} \cite{ricanek2006morph}: The Morph dataset is the largest public available dataset for evaluating age synthesis, age estimation and age invariant verification in the constrained setting, which contains 55,349 color images of 13,672 subjects with age and gender information. The subject ages of Morph range from 16 to 77 years old. It is about four images of each subject on average in the Morph dataset.
\item \textbf{FG-NET} \cite{lanitis2002toward}: The FG-NET dataset is a popular benchmark for age synthesis, age estimation and age invariant verification, which only contains 1,002 images of 82 subjects. The ages of FG-NET range from 0 to 69, while the maximum age gap between images for the same subject is 54 years. We adopt FG-NET as testing set to make fair comparisons with prior works.
\end{itemize}

Five-fold cross-validation is conducted on CACD2000 and Morph. In particular, we divide the dataset into five folds with one fold for testing and the other four folds for training. On CACD2000, each fold contains 400 subjects and the face images in each fold are divided into four age groups: 14-30, 31-40, 41-50, 51-62 with about 9,536, 8,507, 7,890 and 5,959 face images, respectively. On Morph, the face images in each fold are divided into four age groups: 16-30, 31-40, 41-50, 51-77 with about 4,946, 3,207, 2,266 and 616 face images. Note that, for the settings of the two datasets, there are no overlap subjects between the training and testing sets.

\begin{figure*}[t]
\setlength{\abovecaptionskip}{0cm}
\begin{center}
\includegraphics[width=1\linewidth]{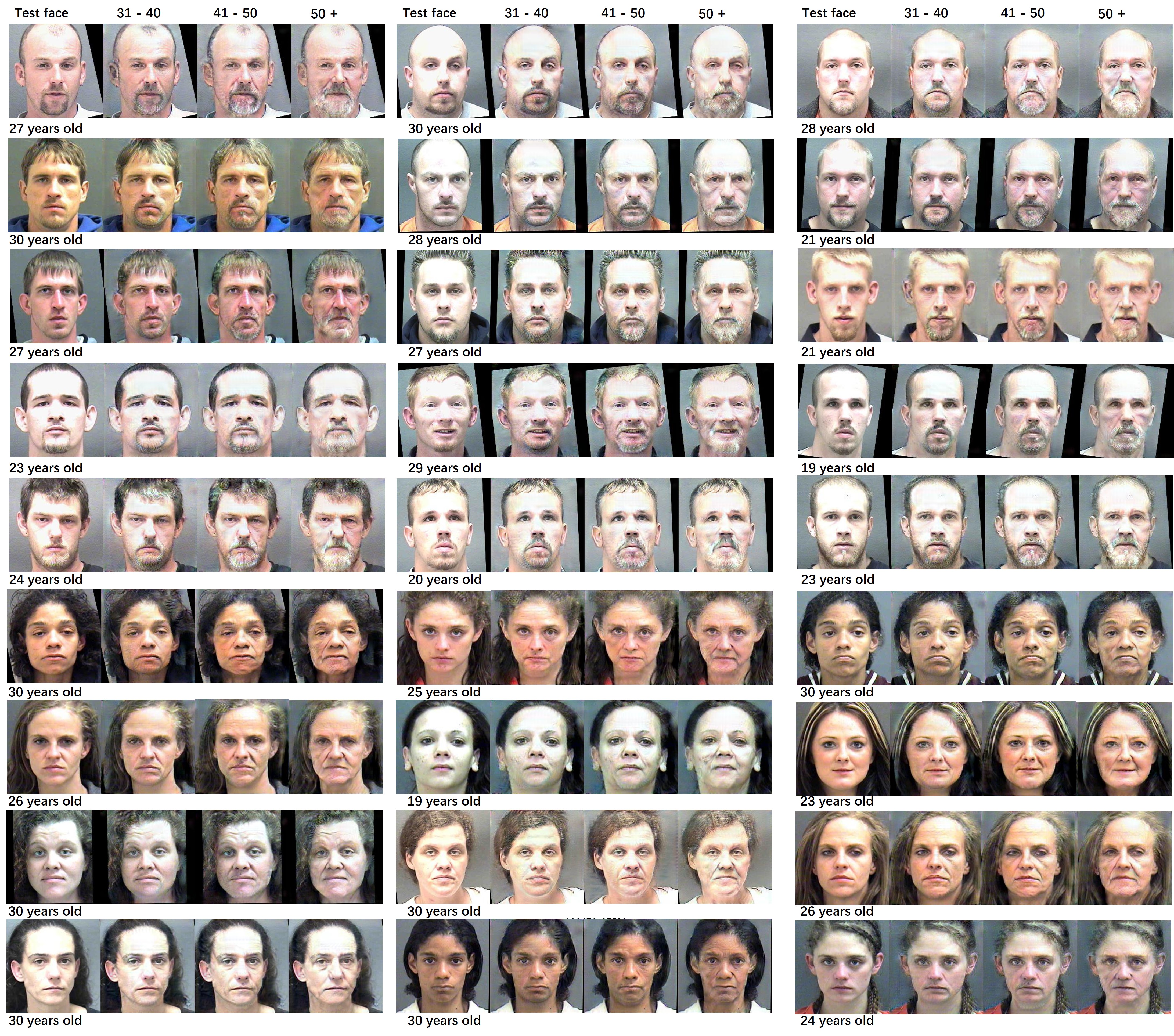}
\end{center}
\vspace{-0.3cm}
   \caption{Age progression results on the Morph dataset for 27 different subjects. For each subject, the leftmost column shows the input face, while the rest three columns are synthetic faces from younger to older. (Zoom in for a better view.)}
\vspace{-0.3cm}
\label{fig:morph_progression}
\end{figure*}
\begin{figure*}[t]
\setlength{\abovecaptionskip}{0cm}
\begin{center}
\includegraphics[width=1\linewidth]{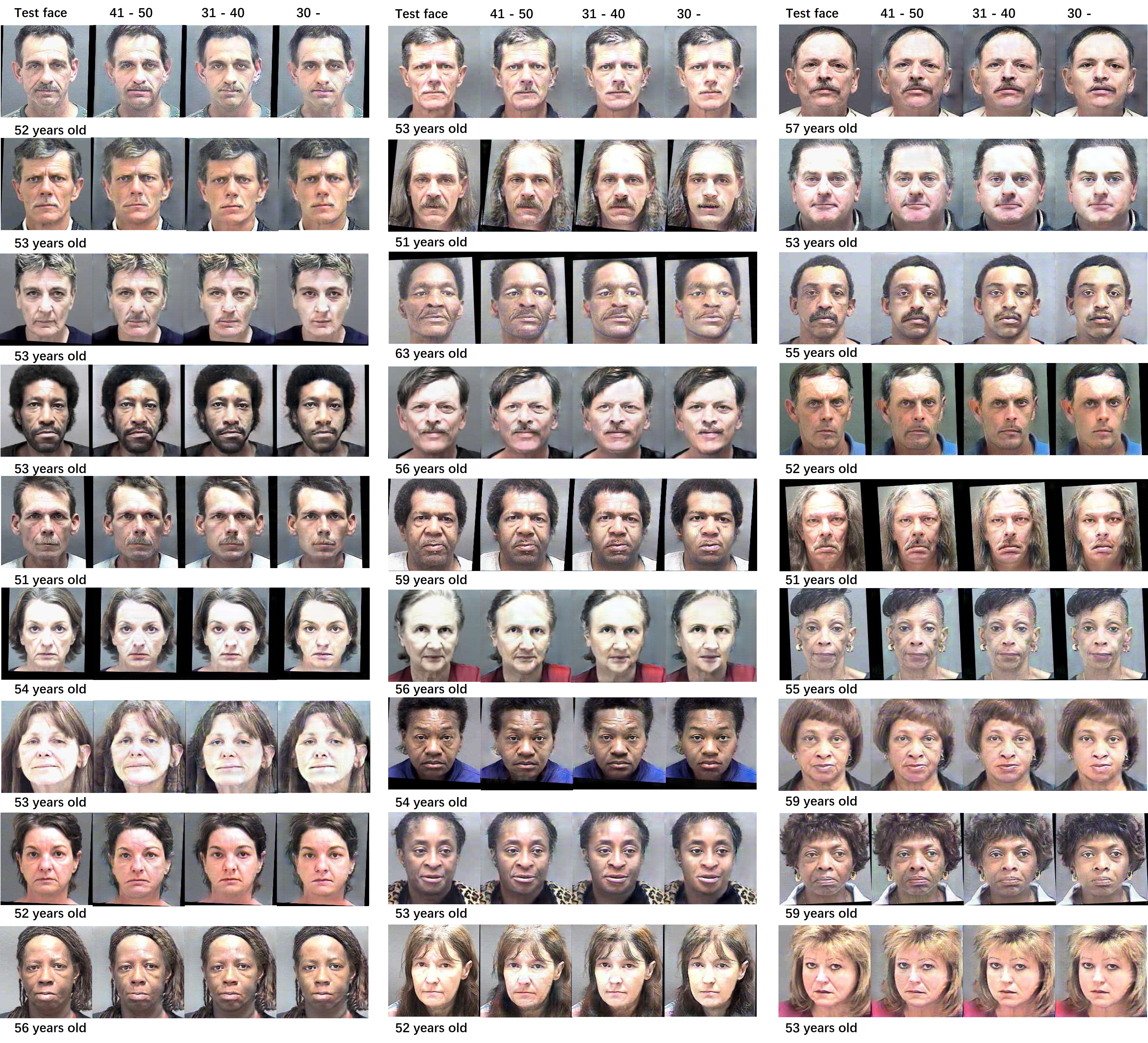}
\end{center}
\vspace{-0.3cm}
   \caption{Age regression results on the Morph dataset for 27 different subjects. For each subject, the leftmost column shows the input face, while the rest three columns are synthetic faces from older to younger. (Zoom in for a better view.)}
\vspace{-0.3cm}
\label{fig:morph_regression}
\end{figure*}

\subsubsection{\textbf{Experimental Settings}}
The identity evaluation network is pretrained on MSCeleb-1M \cite{guo2016ms}. The age evaluation network is pretrained on IMDB-WIKI \cite{rothe2018deep}, while fine-tuned on the corresponding training sets. During training procedure, we fix the parameters of the identity evaluation network and age evaluation network. We utilize multi-task cascaded CNN \cite{zhang2016joint} to detect the bounding box and five facial landmarks of the face images and align the face images by affine transformation to make the two eyes horizontal. According to the detected landmarks, we crop three subregions of forehead, eyes and mouth. All the images are resized to 224 $\times$ 224. Then we convert all the images in the range [0, 255] to the tensors in the range [0.0, 1.0] and send them into the WaveletGLCA-G.
Finally, the output of the WaveletGLCA-G is a tensor in the range [0.0, 1.0]. Before storing as images, the output tensors are multiplied by 255 and taken integers downwardly.

  Our model is implemented with Pytorch. During training, we choose Adam optimizer with ${\beta _1}$ of 0.5, ${\beta _2}$ of 0.99, learning rate of $2 \times {10^{ - 4}}$, and the batch size is set to 8. For CACD2000, the trade-off parameters ${\lambda _1}$, ${\lambda _2}$, ${\lambda _3}$, ${\lambda _4}$ and ${\lambda _5}$ are set to 1.00, 0.01, 120.00, 45.00 and 0.00003, respectively. While for Morph, they are set to 1.00, 0.01, 80.00, 45.00 and 0.0001, respectively. In addition, we update pixel-wise loss at every 10 iteration, and update the discriminator for every 2 generator iterations.

On one GeForce GTX TITAN X and 32 Intel(R) Xeon(R) CPU E5-2660 0 @ 2.20GHz, it takes 72 hours for Morph and 108 hours for CACD2000.

Specifically, the global specific network $G_{g}$ takes the whole image ${I^g} \in {R^{224 \times 224 \times 3}}$ concatenated with the target age label ${L}\in {R^{224 \times 224 \times 4}}$ as its input $\left( {{I^g},L} \right)\in {R^{224 \times 224 \times 7}}$ and its output is $G_{g}(I^g,L) \in {R^{112 \times 112 \times 256}}$.
Simultaneously, the three local specific networks $G_{l}^i,i \in \left\{ {1,2,3} \right\}$ take the forehead, eyes and mouth local patches ${p_i} \in {R^{224 \times 224 \times 3}}$ concatenated with the target age label ${L}\in {R^{224 \times 224 \times 4}}$ as the inputs and learn a separate set of filters to extract and transform the corresponding local feature maps $G_{l}^i(p_i,L) \in {R^{112 \times 112 \times 256}}$.
The feature fusion network concatenates the $G_{g}(I^g,L)$ and three $G_{l}^i(p_i,L)$ as its input and outputs a fused feature $f\in {R^{112 \times 112 \times 512}}$ for wavelet coefficient prediction network. Then the wavelet coefficient prediction network takes $f$ as its input and predicts four specific wavelet coefficients $w_{j},j \in \left\{ {1,2,3,4} \right\}$ by four individual subnetworks, where all the $w_{j}\in {R^{112 \times 112 \times 3}}$.
Finally, the wavelet reconstruction network concatenates the four $w_{j},j \in \left\{ {1,2,3,4} \right\}$ as its input and reconstructs the target age face $\hat I^t\in {R^{224 \times 224 \times 3}}$.

 The age evaluation network takes the synthesized image $\hat I^t\in {R^{224 \times 224 \times 3}}$ as its input. It predicts the age $\hat y$ of $\hat I^t$. The identity evaluation network takes the input image and the synthesized image as its inputs and outputs the face features. Besides, the discriminator takes the input image, the synthesized image and real image in the target age as its inputs, and outputs a scalar, representing the probability that the inputs come from the real data.

\subsection{Qualitative Evaluation of WaveletGLCA-GAN}
In this subsection, we present and compare the visualization results against state-of-the-art age synthesis methods.

\subsubsection{\textbf{Results of Age Progression and Regression}}

WaveletGLCA-GAN can synthesize any age faces controlled by the age labels. Hence, it is able to achieve both age progression and regression. We firstly show the age progression results on CACD2000 and Morph in Fig. \ref{fig:cacd_progression} and Fig. \ref{fig:morph_progression}, respectively. The first column of each subject is the input face under 30 years old. The second, third and fourth columns are the synthetic results of WaveletGLCA-GAN for 31-40, 41-50 and 51-62(77) in turn. We can see that as the age labels increasing, the synthetic results become older and older. Specifically, nasolabial folds and the lines at the sides of the eyes are deepened, white hair and beards emerge and gradually increase. Meanwhile, the age regression results on CACD2000 and Morph are shown in Fig. \ref{fig:cacd_regression} and Fig. \ref{fig:morph_regression}, respectively. The first column of each subject is the input beyond 51 years old. The second, third and fourth columns are the rejuvenating results of WaveletGLCA-GAN for 41-50, 31-40 and 14(16)-30 in turn. Obviously, with the age label decreasing, the white beards and hair are gradually turned into black, dry skin begin to restore elasticity and wrinkles are reduced or even disappear.
\begin{figure}
\centering  
\subfigure[Hair Rejuvenating]{
\label{Hair Rejuvenating}
\includegraphics[width=0.23\textwidth]{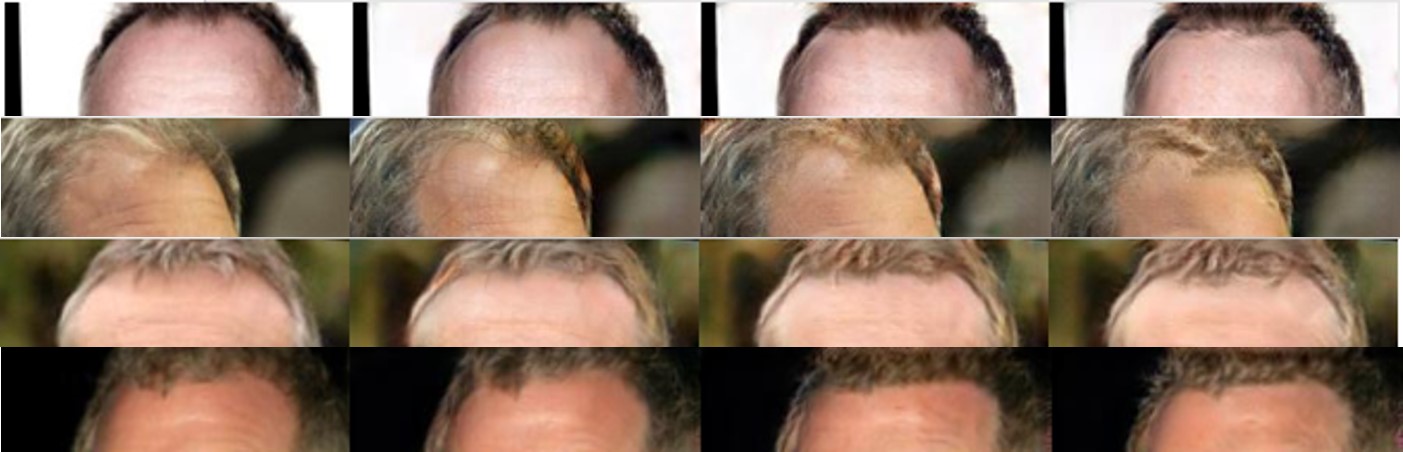}}
\vspace{-0.2cm}
\subfigure[Eyes Rejuvenating]{
\label{Fig.sub.}
\includegraphics[width=0.23\textwidth]{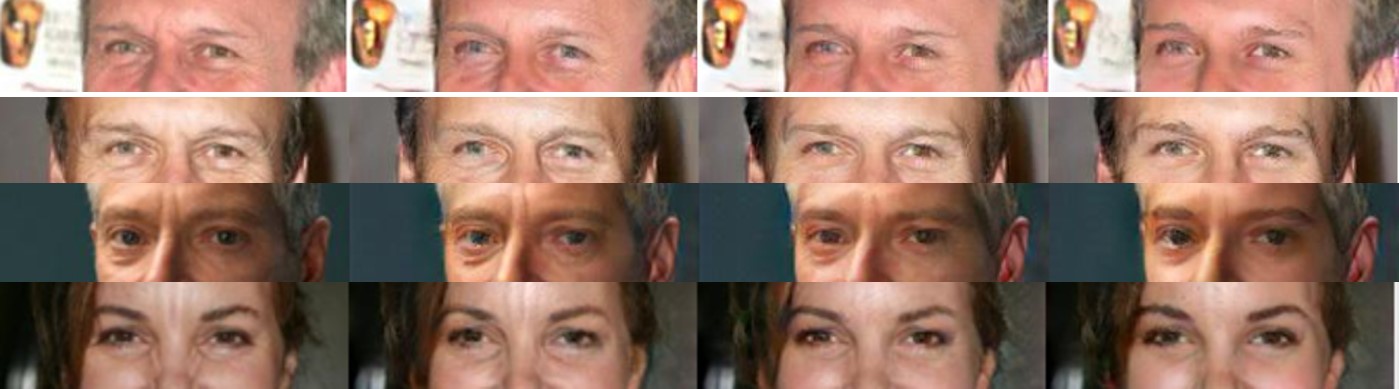}}
\vspace{-0.2cm}
\subfigure[Mouth Rejuvenating]{
\label{Fig.sub.3}
\includegraphics[width=0.23\textwidth]{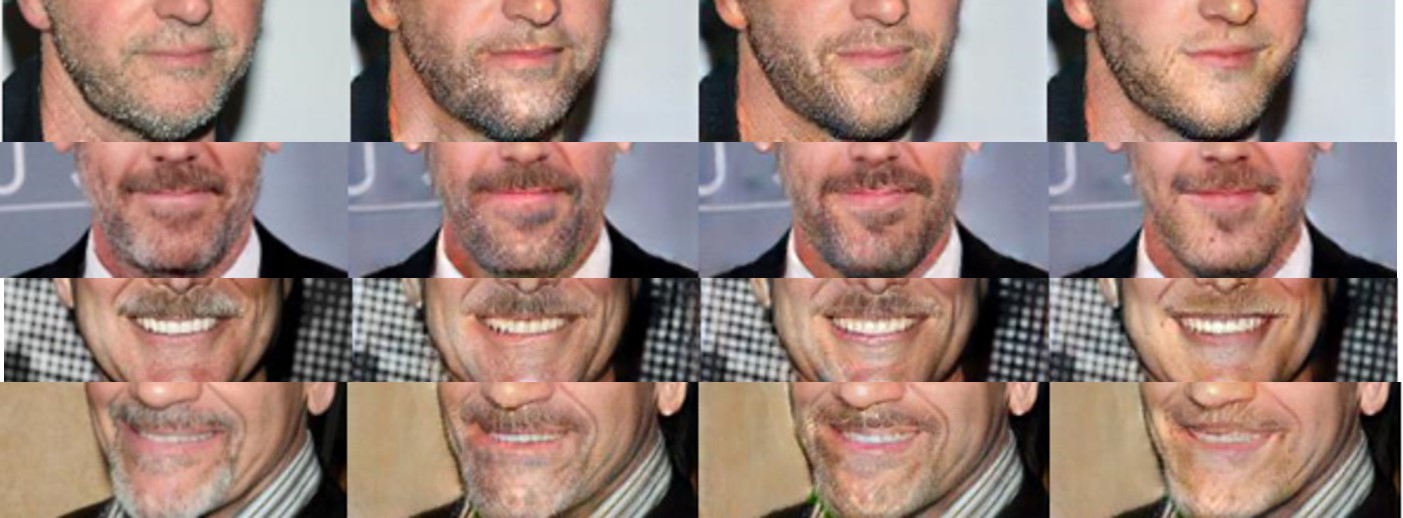}}
\vspace{0.1cm}
\subfigure[Half face Rejuvenating]{
\label{Fig.sub.4}
\includegraphics[width=0.23\textwidth]{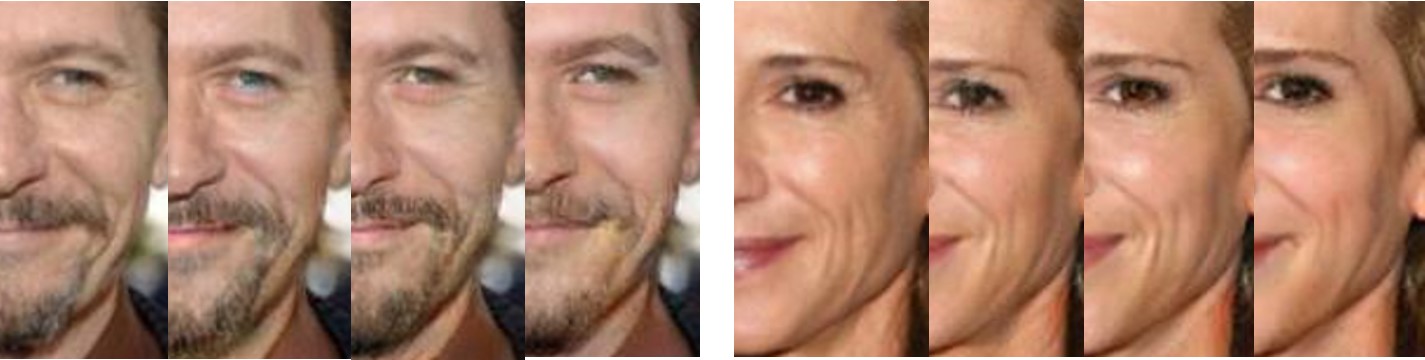}}
\vspace{0.1cm}
\caption{Illustration of visual fidelity. For each subject, the first column is the input, while the rest three columns are the rejuvenating results. (Zoom in for a better view.)}
\label{fig:illustration}
\end{figure}

\begin{figure}[t]
\setlength{\abovecaptionskip}{0cm}
\begin{center}
\includegraphics[width=0.8\linewidth]{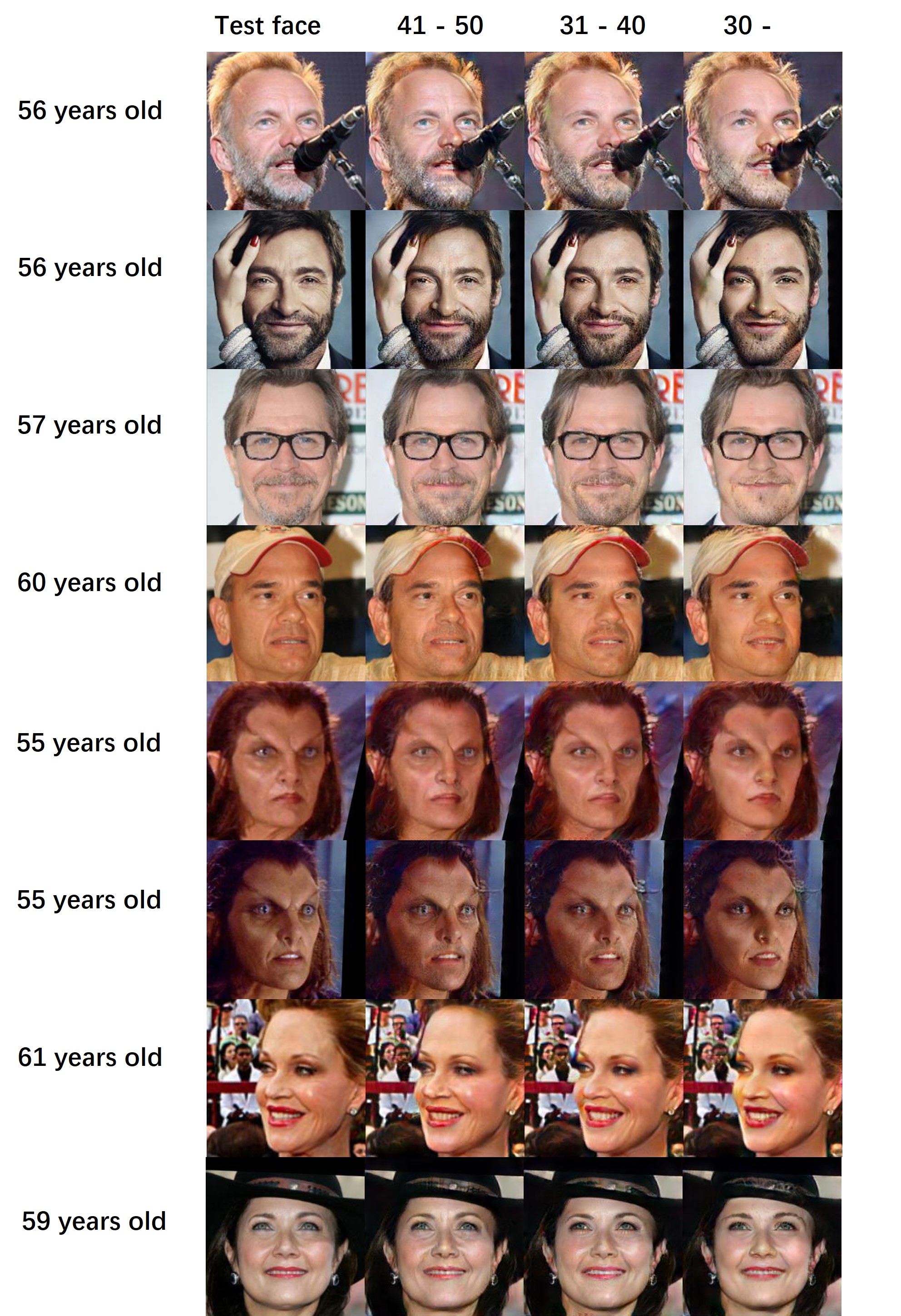}
\end{center}
\vspace{-0.2cm}
   \caption{Robustness to occlusion, glasses, hat, makeup and pose. For each subject, the first column is the input, while the rest three columns are the rejuvenating results. (Zoom in for a better view.) }
\vspace{-0.2cm}
\label{fig:robustness}
\end{figure}

\begin{figure}[t]
\setlength{\abovecaptionskip}{0cm}
\begin{center}
\includegraphics[width=0.7\linewidth]{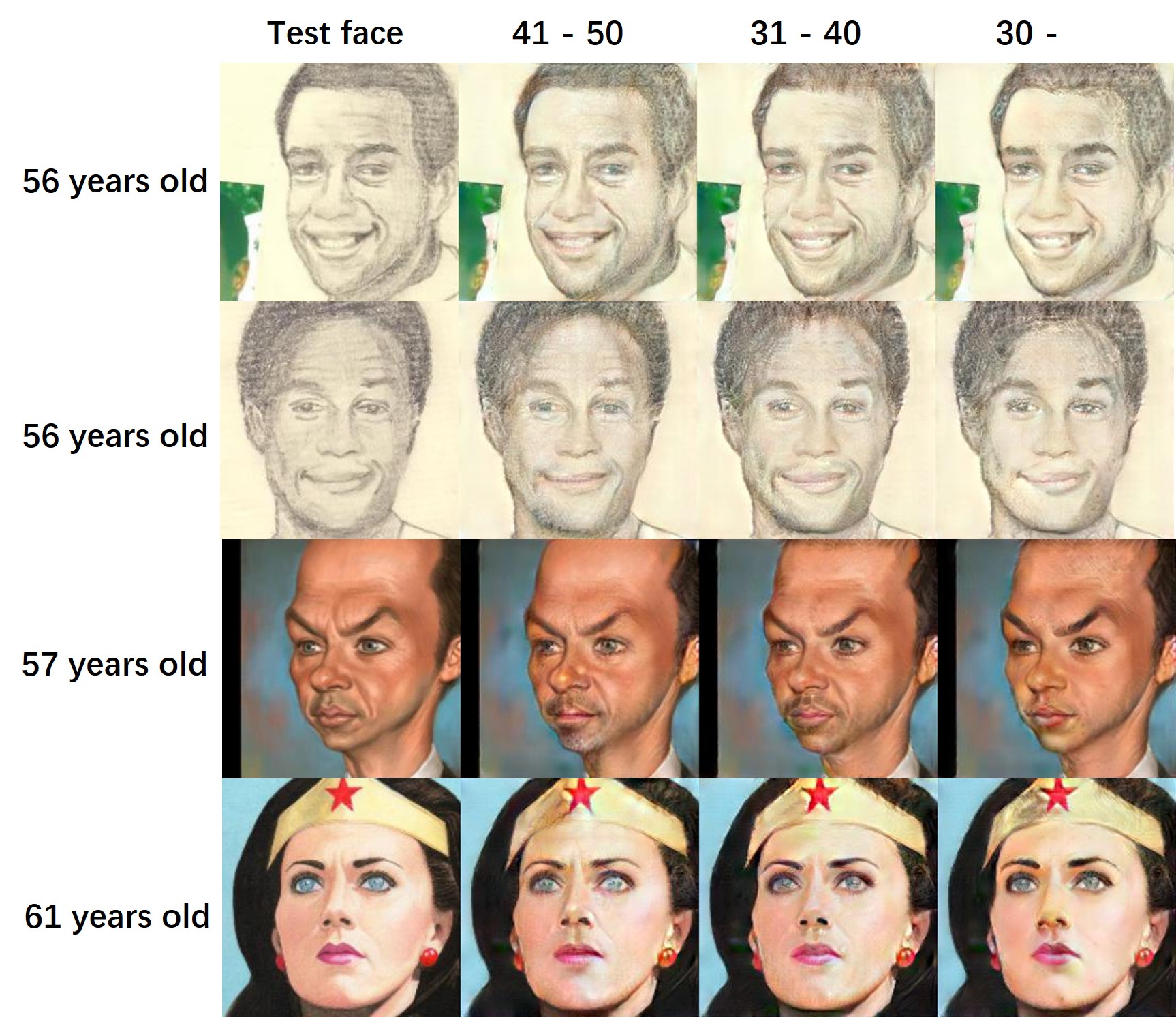}
\end{center}
\vspace{-0.2cm}
   \caption{Generalization to sketch images. For each subject, the first column is the input, while the rest three columns are the rejuvenating results. (Zoom in for a better view.) }\label{tab:sketch}
\vspace{-0.3cm}
\end{figure}

In addition, the changes of the synthetic results are consecutive and consistent with the age label changes. Fig. \ref{fig:illustration} details the changes of synthetic results in the age rejuvenating process. In Fig. \ref{fig:illustration}(a), the hairline gradually becomes lower, the hair darkens and thickens. Fig. \ref{fig:illustration}(b) shows that with the age labels decreasing, the lines at the sides of the eyes get smoother, the eyes get bigger and brighter. In Fig. \ref{fig:illustration}(c), the beards become less and turn black from white. Fig. \ref{fig:illustration}(d) demonstrates the rejuvenating consistency of the WaveletGLCA-GAN. Besides, as is shown in Fig. \ref{fig:robustness}, the WaveletGLCA-GAN is robust to pose, expression and illumination variations. For example, the occlusion, i.e., microphone, glasses, sunglasses, hat and makeup, is also well preserved in the age
rejuvenating process. The proposed WaveletGLCA-GAN also has pretty generalization performance. Fig. \ref{tab:sketch} presents the face rejuvenating results with the sketch face images as the inputs.

\subsubsection{\textbf{Wavelet Packet Decomposition in Age Synthesis}}
\begin{figure*}[t]
\setlength{\abovecaptionskip}{0cm}
\begin{center}
\includegraphics[width=0.8\linewidth]{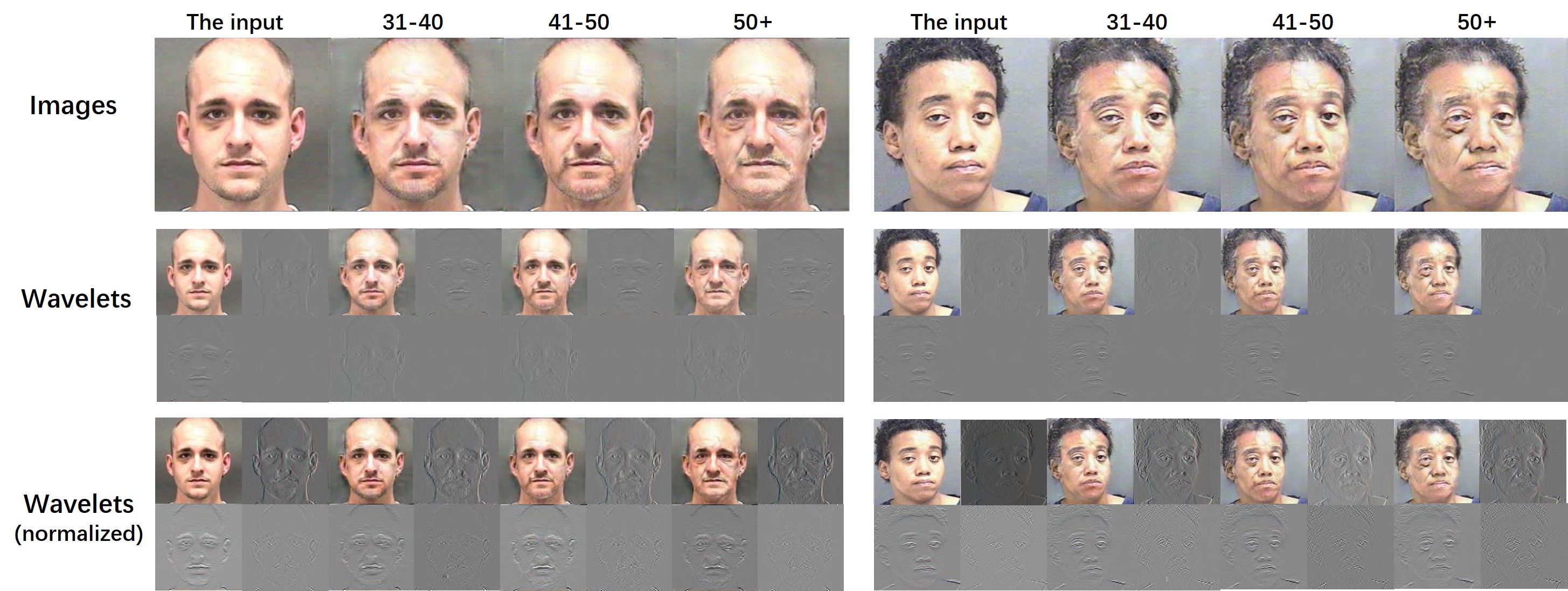}
\end{center}
\vspace{-0.2cm}
   \caption{Illustration of age progression and its corresponding wavelet packet decomposition. Top row: age progression results on the Morph dataset. For each subject, the leftmost column depicts the input face, while the rest three columns are synthetic results from younger to older. The second and bottom rows: the 1-level full wavelet packet decomposition and normalized 1-level full wavelet packet decomposition of the corresponding images in the same columns. (Zoom in for a better view.) }\label{tab:de}
\vspace{-0.3cm}
\end{figure*}

Fig. \ref{tab:de} depicts the age progression results on Morph and its corresponding visual results of wavelet packet decomposition. The low-frequency component is related to the general face information, while the three high-frequency components are related to the detailed aging texture information. We observe that as the age increasing, the texture in both low-frequency and high-frequency components are more abundant, and the hair in the low-frequency component gradually becomes white from black.
\subsubsection{\textbf{Comparison with Prior Works}}

\begin{figure*}[t]
\begin{center}
\includegraphics[width=0.88\linewidth]{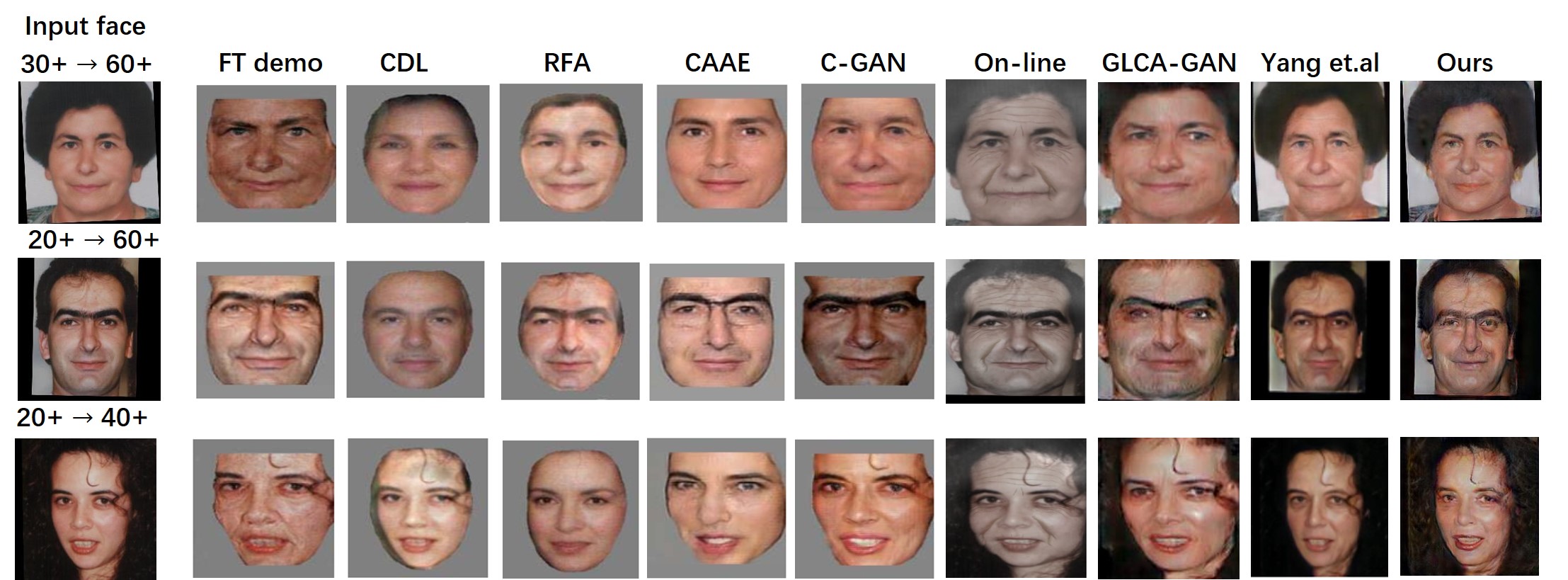}
\end{center}
\vspace{-0.2cm}
   \caption{Comparison with prior works FT demo \cite{FTdemo}, CDL \cite{shu2015personalized}, RFA \cite{wang2016recurrent}, CAAE \cite{zhang2017age}, C-GAN \cite{liu2017face}, AgingBooth App \cite{AgingBooth}, GLCA-GAN \cite{li2018global}, and Yang et.al \cite{yang2017learning}. (Zoom in for a better view.)}\label{Fig.com}
\vspace{-0.2cm}
\end{figure*}

Different from the most previous works, which only focus on the age synthesis of cropped faces, our proposed WaveletGLCA-GAN can produce the entire face images, including complete foreground and background. We compare the aging results of WaveletGLCA-GAN with different methods, including HFA: hidden factor analysis \cite{yang2016face}, FT demo: Face Transformer demo \cite{FTdemo}, CDL: coupled dictionary learning \cite{shu2015personalized}, RFA: recurrent face aging \cite{wang2016recurrent}, CAAE: conditional adversarial autoencoder \cite{zhang2017age}, C-GAN: contextual generative adversarial nets \cite{liu2017face}, GLCA-GAN: global and local consistent age generative adversarial networks \cite{li2018global}, Yang et.al \cite{yang2017learning} as well as an popular mobile aging application, i.e. AgingBooth \cite{AgingBooth}. Fig. \ref{Fig.com} depicts the comparisons. Note that our WaveletGLCA-GAN is trained only on CACD2000 and tested on FG-NET. For fair comparison, we choose the same faces as their papers, and directly cite their synthetic results. As respected, WaveletGLCA-GAN obtains the best visual results with both backgrounds and foregrounds well preserved, while \cite{yang2016face, zhang2017age, FTdemo, shu2015personalized, wang2016recurrent, liu2017face} only generate cropped faces.
Compared with previous best method \cite{yang2017learning}, the synthetic images of our WaveletGLCA-GAN are clearer and have smaller color deviation, which is benefit from the introduce of the residual faces. Besides, compared with Yang et.al \cite{yang2017learning}, our WaveletGLCA-GAN also truly simulates the changes of hairlines, i.e., the hairlines get higher in aging process as shown in Fig. \ref{Fig.com}.

\subsection{Quantitative Evaluation of WaveletGLCA-GAN}
Age accuracy and identity preserving are two underlying requirements for age synthesis. We conduct age estimation and face verification to quantitatively evaluate how much age information and identity information are predicted or preserved.
Specifically, if the generated images can result in better aging accuracy, more precise age information is predicted. If the generated images result in better verification accuracy, more identity information is preserved during aging/rejuvenating process.

\subsubsection{\textbf{Age Estimation}}
Different subjects have different aging processes, but the overall trends of the synthetic faces should be robust and consistent with the real data. Following \cite{yang2017learning}, we apply the online face analysis tool of Face++ \cite{ Face++} to estimate the ages of synthetic faces. Table \ref{tab:agecacd} shows the age estimation results for different age groups on CACD2000. We choose faces under 30 years old (age group0, AG0) as the input test images, synthesize faces in 31-40 (AG1)£¬ 41-50 (AG2) and 51-62 (AG3), and calculate the average age estimated by \cite{ Face++}. As shown in Table \ref{tab:agecacd}, the mean values of the natural faces for 31-40, 41-50 and 51-62 are 39.15, 47.14 and 53.87, respectively, while the mean values of the generated results of WaveletGLCA-GAN are 37.56, 48.13 and 54.17. In addition, we compare WaveletGLCA-GAN with CAAE \cite{zhang2017age}, Yang et.al \cite{yang2017learning} and GLCA-GAN \cite{li2018global} on CACD2000, and observe that the results of WaveletGLCA-GAN are closer to the natural faces.
\begin{table}[htbp]\scriptsize
\centering
\caption{Comparisons of age estimation results (in years) on CACD2000.}

\begin{tabular}{lcccccccc}
\hline
\multirow{1}*{\textbf{Method}}&\multicolumn{1}{c}{\textbf{AG1}}&\multicolumn{1}{c}{\textbf{AG2}}&\multicolumn{1}{c}{\textbf{AG3}}\\
\hline
\hline
CAAE\cite{zhang2017age} &31.32&    34.94 &     36.91\\
Yang et.al\cite{yang2017learning} &44.29 & 48.34  &    52.02\\
GLCA-GAN\cite{li2018global} &37.09 & 44.92  &    48.03\\
\hline

Wavelet-GAN &39.42 & 44.66  &    47.18\\
WaveletGLCA-GAN &37.56 & 48.13  &    54.17\\
\hline

\textbf{Real Data} &39.15 & 47.14  &   53.87\\
\cline{1-4}
\end{tabular}\label{tab:agecacd}
\end{table}

Meanwhile, Table \ref{tab:morph} presents the age estimation results of different methods on Morph. We choose faces under 30 years old (age group0, AG0) as the input test images, synthesize faces in 31-40 (AG1)£¬ 41-50 (AG2) and 51-77 (AG3), and calculate the average age estimated by [40]. The mean values of the natural faces for 31-40, 41-50 and 51-67 are 38.59, 48.24 and 58.28 respectively, while the mean values of the synthetic results of WaveletGLCA-GAN are 38.36, 46.90 and 59.14. In addition, we compare WaveletGLCA-GAN with CAAE \cite{zhang2017age}, Yang et.al \cite{yang2017learning} and GLCA-GAN \cite{li2018global} on Morph, and observe that the age accuracy of WaveletGLCA-GAN outperforms other methods.
\begin{table}[htbp]\scriptsize
\centering
\caption{Comparison of age estimation results (in years) on Morph.}

\begin{tabular}{lcccccccccc}
\hline
\multirow{1}*{\textbf{Method}}&\multicolumn{1}{c}{\textbf{AG1}}&\multicolumn{1}{c}{\textbf{AG2}}&\multicolumn{1}{c}{\textbf{AG3}}\\
\hline
\hline
CAAE\cite{zhang2017age} &28.13&32.50 &36.83\\
Yang et.al\cite{yang2017learning} &42.84 & 50.78  &    59.91\\
GLCA-GAN\cite{li2018global} &43.00 & 49.03  &    54.60\\
\cline{1-4}

Wavelet-GAN &44.62 & 49.92  &    55.17\\
WaveletGLCA-GAN &38.36 & 46.90  &    59.14\\
\cline{1-4}
\textbf{Real Data} &38.59 & 48.24  &    58.28\\
\cline{1-4}
\end{tabular}\label{tab:morph}
\end{table}
\subsubsection{\textbf{Face Verification}}



\begin{table*}[htbp]\scriptsize
\centering
\caption{Face verification results($\%$) on CACD2000 and Morph.}

\begin{tabular}{llllllll}
\hline
\multirow{2}{*}{\textbf{Progression}}&\multicolumn{1}{l}{CACD2000}&\multicolumn{1}{l}{Morph}&\multicolumn{1}{c}{            }&\multirow{2}{*}{\textbf{Regression}}&\multicolumn{1}{l}{CACD2000}&\multicolumn{1}{l}{Morph}\\
\cline{2-3}
\cline{6-7}

&\multicolumn{1}{c}{TAR@FAR=$10^{-5}$}&\multicolumn{1}{c}{TAR@FAR=$10^{-5}$}&&&\multicolumn{1}{c}{TAR@FAR=$10^{-5}$}&\multicolumn{1}{c}{TAR@FAR=10$^{-4}$}\\
\hline
Test Face$\rightarrow$AG1&97.71&99.94 &$\quad$&Test Face$\rightarrow$AG2 &98.85&100 \\
Test Face$\rightarrow$AG2 &96.07& 99.92&$\quad$ &Test Face$\rightarrow$AG1&98.91&100 \\
Test Face$\rightarrow$AG3  &95.25&99.12  & $\quad$&Test Face$\rightarrow$AG0&98.53&99.85  \\
AG1$\rightarrow$AG2 &99.22& 99.94&$\quad$&AG2$\rightarrow$AG1&99.62&100 \\
AG1$\rightarrow$AG3 &96.01& 99.68 &$\quad$&AG2$\rightarrow$AG0&98.06& 99.70 \\
AG2$\rightarrow$AG3 &97.59 & 99.86 &$\quad$&AG1$\rightarrow$AG0&99.00&100 \\

\hline
\end{tabular}\label{tab:verification}
\end{table*}

We evaluate the identity preserving performance of WaveletGLCA-GAN by face verification for both age progression and regression. In the age progression setting, for each testing face, we evaluate the verification rate between the input image and its corresponding age synthetic result: [test face$\rightarrow$AG1], [test face$\rightarrow$AG2] and [test face$\rightarrow$AG3]. Following \cite{yang2017learning}, we also conduct face verification among the synthetic faces: [AG1$\rightarrow$AG2], [AG1$\rightarrow$AG3] and [AG2$\rightarrow$AG3]. Similar with age progression, in the age regression, we evaluate on [test face$\rightarrow$AG2], [test face$\rightarrow$AG1], [test face$\rightarrow$AG0], [AG2$\rightarrow$AG1], [AG2$\rightarrow$AG0] and [AG1$\rightarrow$AG0]. For CACD2000, we randomly choose negative pairs, and make ${\rm{positive pairs:negative pairs = 1:100}}$. For Morph,  we randomly choose negative pairs, and make ${\rm{positive pairs:negative pairs = 1:200}}$, due to the lack of old age images (about 616 images of 51-77 in each fold).
Light CNN \cite{wu2015light} is employed as the feature extractor in our experiments. We report TAR@FAR=$10^{-5}$ on CACD2000. While due to the lack of old age images in Morph, we only report TAR@FAR=$10^{-5}$ for age progression and TAR@FAR=$10^{-4}$ for age regression. Table \ref{tab:verification} presents the face verification results.

\subsection{Ablation Study}
\subsubsection{\textbf{Contribution of Local Specific Networks}}

To prove the effectiveness of the proposed three local specific networks, we compare the age accuracy  of WaveletGLCA-GAN with Wavelet-GAN (The model only use global specific network to synthesize target age faces and keep others consistent with WaveletGLCA-GAN).
As Tables \ref{tab:agecacd} and \ref{tab:morph} shown, WaveletGLCA-GAN is superior to  Wavelet-GAN in age accuracy.

\begin{figure}[t]
\setlength{\abovecaptionskip}{0cm}
\begin{center}
\includegraphics[width=0.8\linewidth]{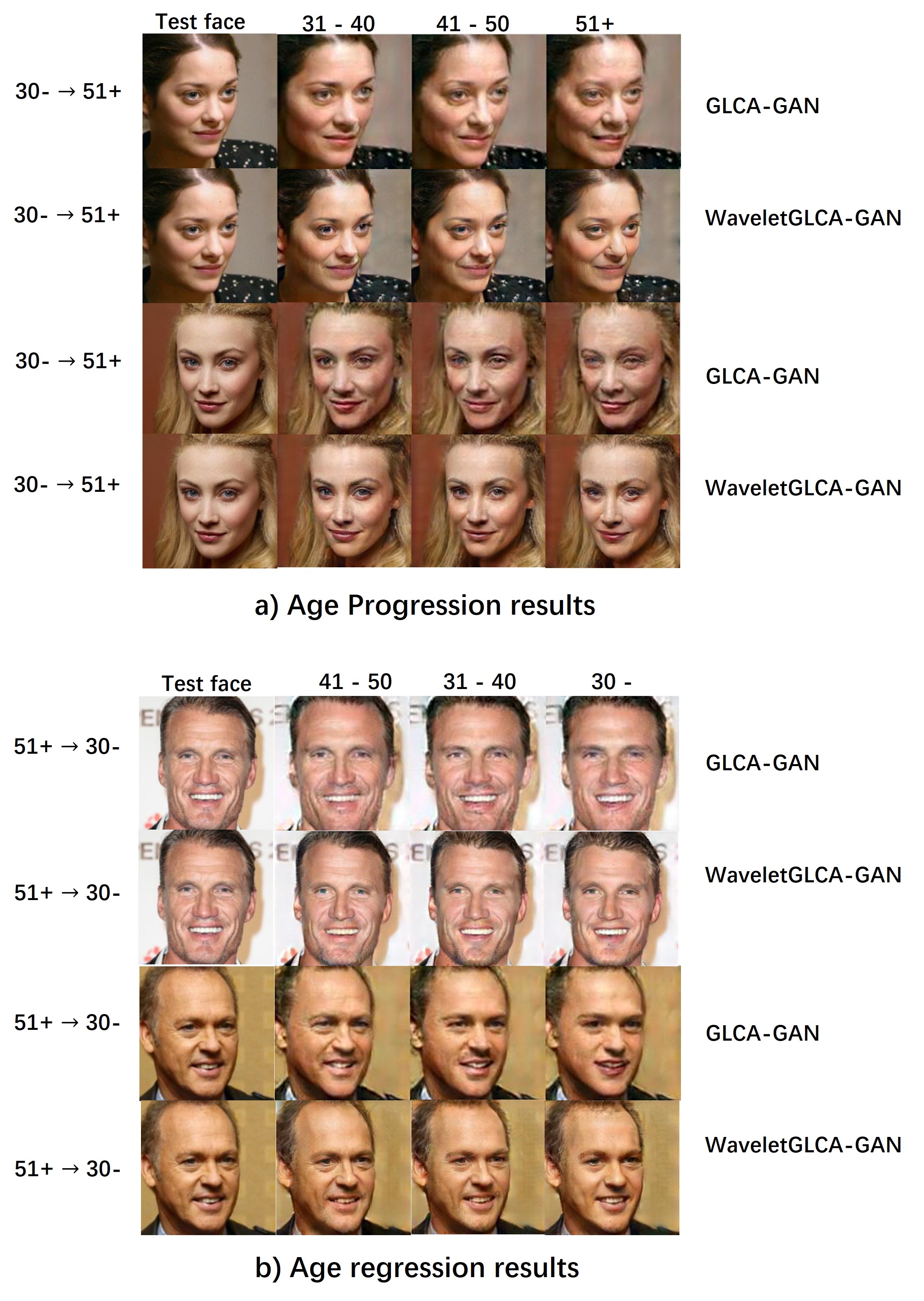}
\end{center}
\vspace{-0.2cm}
   \caption{Comparison to GLCA-GAN (without wavelet coefficient prediction net) on the CACD2000. (Zoom in for a better view.) } \label{Fig.wave}
\vspace{-0.2cm}
\end{figure}

\subsubsection{\textbf{Contribution of Wavelet Coefficient Prediction Net}}

\begin{figure*}[t]
\setlength{\abovecaptionskip}{0cm}
\begin{center}
\includegraphics[width=0.7\linewidth]{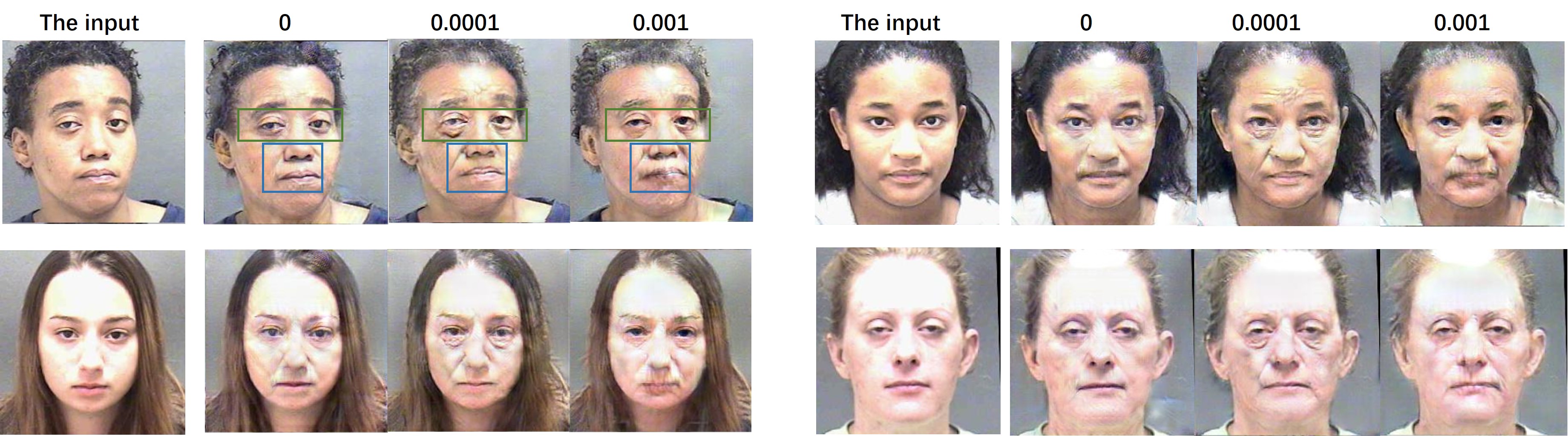}
\end{center}
\vspace{-0.2cm}
   \caption{Aging results on Morph with different trade-off parameters of TV loss, including 0, 0.0001 and 0.001. The first column of each subject is the input image, while the rest columns are the age progression results with different trade-off parameters. (Zoom in for a better view.) } \label{Fig.tvloss}
\vspace{-0.2cm}
\end{figure*}
One assumption of WaveletGLCA-GAN is that the introduced of wavelet transform advances the generation of more subtle age-related texture information. Fig. \ref{Fig.wave} presents the visual comparisons between WaveletGLCA-GAN and GLCA-GAN \cite{li2018global}, which is without wavelet coefficient prediction net. We observe that the synthetic faces by WaveletGLCA-GAN are more photo-realistic. With wavelet coefficient prediction net, more detailed and clear texture information is produced. Furthermore, Tables \ref{tab:agecacd} and \ref{tab:morph} show WaveletGLCA-GAN outperforms GLCA-GAN in age accuracy.
\subsubsection{\textbf{Contribution of TV Loss}}

We evaluate our method on Morph with different trade-off parameters of TV loss, including 0, 0.0001 and 0.001, and present the results in Fig. \ref{Fig.tvloss}.  The first column of each subject is the input image and the rest three columns are the aging results of 50+ years old.
When the trade-off parameter is set to 0, the texture and the hair/beard coloration of the aging results are not nature and have significant effect on the reality. When the trade-off parameter is set to 0.0001, the aging texture looks more realistic as outlined in the green box. Finally, we increase the trade-off parameter to 0.001. Both the artifacts and some aging textures are reduced, and the synthesized results are smoothed over. Thus, in this paper, we choose 0.0001 as the trade-off parameter of TV loss.

\subsection{\textbf{Limitation and Discussion}}
\begin{figure}[t]
\setlength{\abovecaptionskip}{0cm}
\begin{center}
\includegraphics[width=0.7\linewidth]{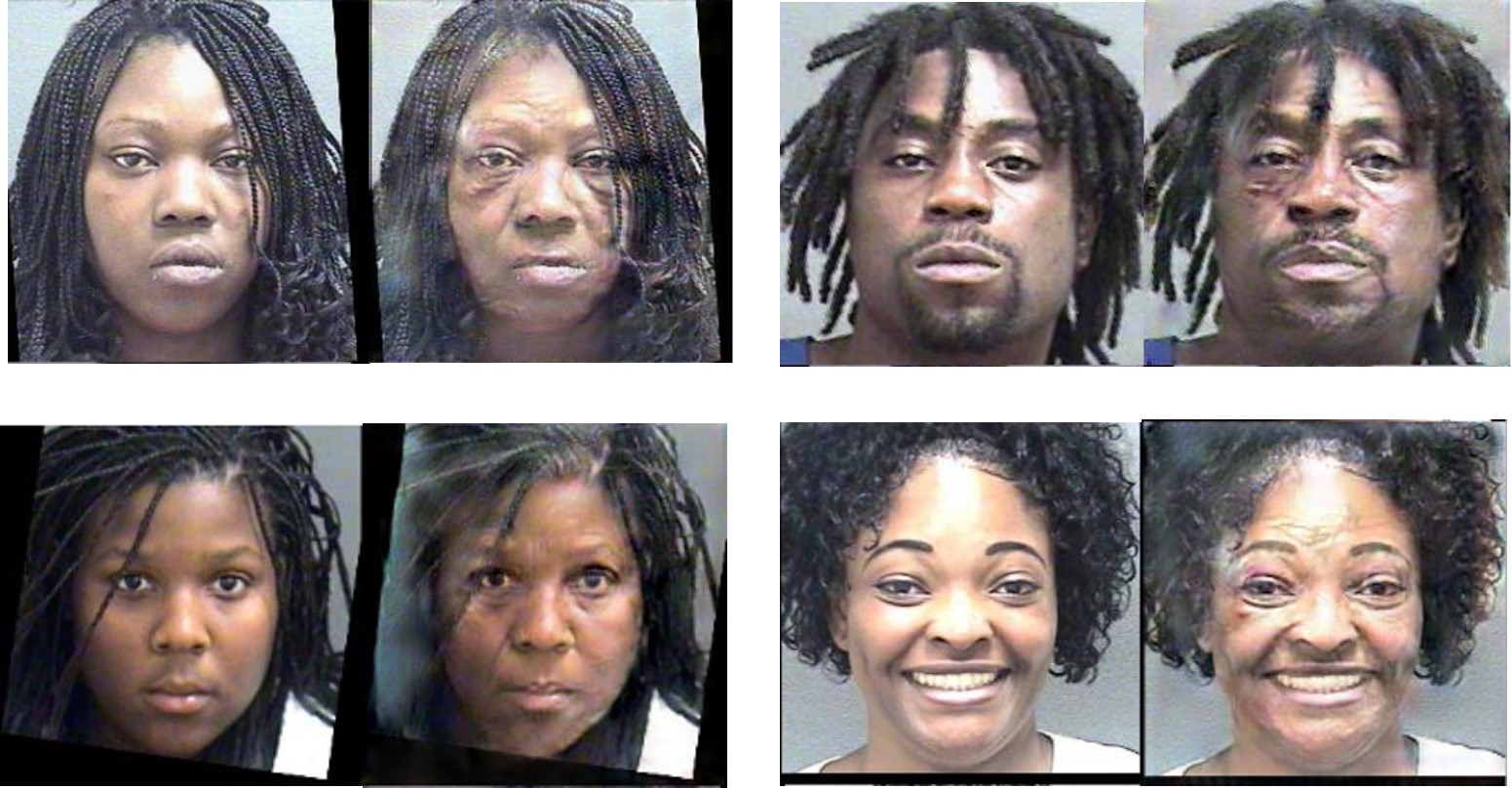}
\end{center}
\vspace{-0.2cm}
   \caption{Typical failure case of our method. For each image pair, the left shows the input face under 30 years old, and the right is the synthetic result beyond 50 years old. (Zoom in for a better view.)} \label{Fig.failure}
\vspace{-0.2cm}
\end{figure}

Although our method achieves photo-realistic age progression/regression results, there is room for improvement. The hair color can be modified on the Morph database, while cannot on the CACD2000 database. Because the CACD2000 is only composed of images of European and American celebrities, whose faces usually look younger than their actual ages. Besides in CACD2000, many facial images under 20 years old have white-blonde hair, and there are also many facial images beyond 50 years old have colorful instead of white hair, which leads to an extreme difficulty for the model learning to change the hair color. Since the data in Morph is clean and our model has the ability to simulate the color changes of hair and beards. However, if the input face has thick black long hair as shown in Fig. \ref{Fig.failure}, there sometimes appears blurry and white irregular areas in the age progression/regression results. The change of hair color is important to face age progression/regression process, thus we will continue to explore this in the further work.

Besides, most of the current age synthesis models mainly focus on the aging or rejuvenating between adult faces. However, it is equally important and more challenging to explore the aging or rejuvenating between adult and baby faces. In the future, we intend to model the continuous aging mechanism among all the ages.

\section{Conclusion}
This paper proposes a Wavelet-domain Global and Local Consistent Age Generative Adversarial Network (WaveletGLCA-GAN) method to synthesize a face conditioned on the age labels. WaveletGLCA-GAN simultaneously achieves age progression and regression with the given age labels and generates favorable results. With one global specific network and three local specific networks, both global topology information and local texture details of the input faces can be captured. By introducing the frequency domain information, the synthesized results are clearer and more sensitive to facial texture. Five types of losses are adopted to supervise the synthesizer to achieve accurate age generation under the premise of preserving the identity information. Extensive experimental results on age synthesis, age estimation and face verification demonstrate that the flexibility, generality and efficiency of the proposed WaveletGLCA-GAN.

\section{ACKNOWLEDGMENT}
This work is funded by State Key Development Program (Grant No. 2017YFC0821602, 2016YFB1001001, 2016YFB1001000), the National Natural Science Foundation of China (Grants No. 61622310, 61427811, 61573360), and Beijing Natural Science Foundation (Grants No. JQ18017).


%

\appendices




\ifCLASSOPTIONcaptionsoff
  \newpage
\fi



%


\bibliographystyle{IEEEtran}
\bibliography{IEEEabrv,latex12}
%








\end{document}